\documentclass[journal]{IEEEtran}
\usepackage[numbers,sort&compress]{natbib}
\usepackage{times}
\usepackage{epsfig}
\usepackage{graphicx}
\usepackage{amsmath}
\usepackage{amssymb}
\usepackage{algorithm}
\usepackage{algorithmic}
\usepackage{bm}

\usepackage{times}
\usepackage{epsfig}
\usepackage{graphicx}
\usepackage{amsmath}
\usepackage{amssymb}
\usepackage{graphicx}
\usepackage{epstopdf}
\usepackage{subfig}
\usepackage{makecell}
\usepackage{multirow}
\usepackage{array}
\usepackage{bm}
\usepackage{color}
\usepackage{caption}

\newsavebox{\mybox}
\newcolumntype{X}[1]{>{\begin{lrbox}{\mybox}}c<{\end{lrbox}\makecell[#1]{\fbox{\usebox\mybox}}}}

\ifCLASSINFOpdf

\else

\fi

\begin{document}

\title{High Performance Visual Object Tracking with Unified Convolutional Networks}
\author{Zheng Zhu,~\IEEEmembership{Student Member,~IEEE,}
        Wei Zou,
        Guan Huang,
        Dalong Du,
        and Chang Huang% <-this % stops a space
\thanks{Zheng Zhu and Wei Zou are with Institute of Automation, Chinese Academy of Sciences, Beijing, China.

Guan Huang, Dalong Du and Chang Huang are with Horizon Robotics, Inc, Beijing, China.

Zheng Zhu is also with University of Chinese Academy of Sciences, Beijing, China.

Manuscript received June 8, 2018}}

\maketitle

% As a general rule, do not put math, special symbols or citations
% in the abstract or keywords.
\begin{abstract}
Convolutional neural networks (CNN) based tracking approaches have shown favorable performance in recent benchmarks. Nonetheless, the chosen CNN features are always pre-trained in different tasks and individual components in tracking systems are learned separately, thus the achieved tracking performance may be suboptimal. Besides, most of these trackers are not designed towards real-time applications because of their time-consuming feature extraction and complex optimization details. In this paper, we propose an end-to-end framework to learn the convolutional features and perform the tracking process simultaneously, namely, a unified convolutional tracker (UCT). Specifically, the UCT treats feature extractor and tracking process both as convolution operation and trains them jointly, which enables learned CNN features are tightly coupled with tracking process. During online tracking, an efficient model updating method is proposed by introducing peak-versus-noise ratio (PNR) criterion, and scale changes are handled efficiently by incorporating a scale branch into network. Experiments are performed on four challenging tracking datasets: OTB2013, OTB2015, VOT2015 and VOT2016. Our method achieves leading performance on these benchmarks while maintaining beyond real-time speed.
%The standard UCT and UCT-lite can track generic objects at 58 FPS and 154 FPS without further optimization, respectively.
\end{abstract}

% Note that keywords are not normally used for peerreview papers.
\begin{IEEEkeywords}
visual tracking, real-time tracker, convolutional neural networks
\end{IEEEkeywords}

\IEEEpeerreviewmaketitle

\section{Introduction}

Visual object tracking, which tracks a specified target in a changing video sequence automatically, is a fundamental problem in many aspects such as visual analysis \cite{sparseRepresentation}, automatic driving \cite{ma2017research}, pose tracking \cite{zhang2019fastpose,zhang2019exploiting,li2019state} and robotics \cite{maselection, kangadaptive,zhu2018velocity,wang2017motion,huang2018optical,huang2018optical2,
huang2018efficient,huang2019motion,zhu2016std}. On the one hand, a core problem of tracking is how to detect and locate the object accurately in changing scenarios such as illumination variations, scale variations, occlusions, shape deformation, and camera motion \cite{OTB2015,smeulders2014visual,kristan2018sixth,matej2015visual}. On the other hand, tracking is a time-critical problem because it is always performed in each frame of sequences. Therefore, how to improve accuracy, robustness and efficiency are main development directions of the recent tracking approaches.

As a core component of trackers, appearance model can be divided into generative models and discriminative models. In generative models, candidates are searched to minimize reconstruction errors. Representative sparse coding \cite{wang2013online,wang2015inverse} have been exploited for visual tracking. In discriminative models, tracking is regarded as a classification problem by separating foreground and background. Numerous classifiers have been adopted for object tracking, such as structured support vector machine (SVM) \cite{Struck}, boosting \cite{boosting} and online multiple instance learning \cite{MIL}. Recently, significant attention has been paid to discriminative correlation filters (DCF) based methods \cite{KCF,SRDCF,LCT,SAMF} for real-time visual tracking. The DCF trackers can efficiently train a repressor by exploiting the properties of circular correlation and performing operations in the Fourier domain. Thus conventional DCF trackers can perform at more than 100 FPS \cite{KCF,CSK}, and these approaches are significant for real-time applications.  Many improvements for DCF tracking approaches have also been proposed, such as SAMF \cite{SAMF} and fDSST \cite{fDSST} for scale changes, LCT \cite{LCT} for long-term tracking, SRDCF \cite{SRDCF} and BACF \cite{BACF} to mitigate boundary effects. The better performance is obtained but the high-speed property of DCF is broken. Moreover, all these methods use handcrafted features, which hinder their accuracy and robustness.

Inspired by the success of CNN in object classification \cite{AlexNet,ResNet}, detection \cite{FasterRCNN} and segmentation \cite{semanticsegmentation}, the visual tracking community has started to focus on the deep trackers that exploit the strength of CNN in recent years. These deep trackers come from two aspects: one is DCF framework with deep features, which means replacing the handcrafted features with CNN features in DCF trackers \cite{HCF,HDT,CCOT,DeepSRDCF}. The other kind of deep trackers is to design the tracking networks and pre-train them which aim to learn the target-specific features for each new video \cite{SiamFC,FCNT}. Despite their notable performance, most of these approaches separate tracking system into some individual components, which may lead to suboptimal performance. Furthermore, most of trackers are not designed towards real-time applications because of their time-consuming feature extraction and complex optimization details. For example, the speed of winners in VOT2015 \cite{matej2015visual} and VOT2016 \cite{VOT2016} are less than 1 FPS on modern GPUs.

We address these two problems (not end-to-end training and low speed) by introducing a unified convolutional tracker (UCT) to learn the features and perform the tracking process simultaneously. UCT is an end-to-end and extensible framework for tracking, which achieves high performance in terms of both accuracy and speed. Specifically, The proposed UCT treats feature extractor and tracking process both as convolution operation, resulting in a fully convolutional network architecture. In online tracking, the whole patch can be predicted using the foreground response map by one-pass forward propagation. Additionally, efficient model updating and scale handling skills are proposed to ensure tracker's real-time property.
\subsection{Contributions}
The contributions of this paper can be summarized in three folds as follows:

1, We propose unified convolutional networks to learn the convolutional features and perform the tracking process simultaneously. The feature extractor and tracking process are both treated as convolution operation that can be trained simultaneously. End-to-end training enables learned CNN features are tightly coupled with tracking process.

2, In online tracking, efficient updating and scale handling strategies are incorporated into the tracking framework. The proposed standard UCT (with VGG-Net) and UCT-lite (with ZF-Net) can track generic objects at 58 FPS and 154 FPS respectively, which is far beyond real time.

3, Extensive experiments are carried out on four tracking benchmarks: OTB2013, OTB2015, VOT2015 and VOT2016. Results demonstrate that the proposed tracking algorithm performs favorably against existing state-of-the-art methods in terms of accuracy and speed.

This paper extends our work~\cite{UCT}, which is published on ICCV2017 VOT Workshop. In this paper, we additionally perform a comprehensive summary of  unified convolutional networks for high performance visual tracking. Furthermore, we modify the backbone network from ResNet to VGG-Net, and improve the training strategies and training data. These enhancements allow us to increase the performance by fine-tuning more convolutional layers, with higher running speed. The proposed improvements result in superior tracking performance and a speedup. The experiments are extended by evaluating our approach comparing with more state-of-the-art trackers. Finally, we also present results on the VOT2016 benchmark.

The rest of this paper is organized as follows: Section \uppercase\expandafter{\romannumeral2} summarizes related works about recent visual tracking. Unified convolutional networks for tracking is introduced in Section \uppercase\expandafter{\romannumeral3}, including overall architecture, formulation and training process. Section \uppercase\expandafter{\romannumeral4} introduces our online tracking algorithm which consists of model updating and scale estimation. Experiments on four challenging benchmarks are shown in Section \uppercase\expandafter{\romannumeral5}. Section \uppercase\expandafter{\romannumeral6} concludes the paper with a summary.

%-------------------------------------------------------------------------
\section{Related work}
Visual tracking is a significant problem in computer vision systems and a series of approaches have been successfully proposed. Since our main contribution is a UCT framework for high performance visual tracking, we give a brief review on three directions closely related to this work: CNN-based trackers, real-time trackers, and fully convolutional networks (FCN).

\subsection{On CNN-based trackers}

Inspired by the success of CNN in object recognition \cite{AlexNet,ResNet,FasterRCNN,zhu2018two,li2018attention,zhu2018action,zhu2017learning}, researchers in tracking community have started to focus on the deep trackers that exploit the strength of CNN \cite{wang2015transferring, Deeptrack,bai2018multi}. Since DCF provides an excellent framework for recent tracking research, the first trend is the combination of DCF framework with CNN features. In HCF \cite{HCF} and HDT \cite{HDT}, the CNN is employed to extract features instead of handcrafted features, and final tracking results are obtained by combining hierarchical response and hedging weak trackers, respectively. DeepSRDCF \cite{DeepSRDCF} exploits shallow CNN features in a spatially regularized DCF framework. Another trend in deep trackers is to design the tracking networks and pre-train them which aim to learn the target-specific features and handle the challenges for each new video. MDNet \cite{MDNet} trains a small-scale network by multi-domain methods, thus separating domain independent information from domain-specific layers. CCOT \cite{CCOT} employs the implicit interpolation method to solve the learning problem in the continuous spatial domain. These trackers have two major drawbacks: Firstly, they can only tune the hyper-parameters heuristically since feature extraction and tracking process are separate, which are not end-to-end trained. Secondly, most of these trackers are not designed towards real-time applications.

\subsection{On real-time trackers}

Other than accuracy and robustness, the speed of the visual tracker is a crucial factor in many real world applications. Therefore, a practical tracking approach should be accurate and robust while operating at real time. Classical real-time trackers, such as NCC \cite{NCC} and Mean-shift \cite{MeanShift}, perform tracking using matching. Recently, discriminative correlation filters (DCF) based methods, which efficiently train a repressor by exploiting the properties of circular correlation and performing the operations in the Fourier domain, have drawn attentions for real-time visual tracking. Conventional DCF trackers such as MOSSE, CSK and KCF can perform at more than 100 FPS \cite{MOSSE, CSK, KCF}. Subsequently, a series of trackers that follow DCF method are proposed. In fDSST algorithm \cite{fDSST}, the tracker searches over the scale space for correlation filters to handle the variation of object size. Staple \cite{Staple} combines complementary template and color cues in a ridge regression framework. CFLB \cite{CFLB} and BACF \cite{BACF} mitigate the boundary effects of DCF in the Fourier domain. Nevertheless, all these DCF-based trackers employ handcrafted features that hinders their performance.

The recent years have witnessed significant advances of CNN-based real-time tracking approaches \cite{SiamFC,GOTURN,CFNet,DCFNet,zhu2018end,zhu2018end2,li2018high,zhu2018distractor}. SiamFC \cite{SiamFC} proposes a fully convolutional Siamese network to predict motion between two frames. The network is trained off-line and evaluated without any fine-tuning. Similarly to SiamFC, in GOTURN tracker \cite{GOTURN}, the   motion between successive frames is predicted using a deep regression network. These two trackers are able to perform at 86 FPS and 100 FPS respectively on GPU because no fine-tuning is performed online. Although their simplicity and fixed-model nature lead to high speed, this method loses the ability to update the appearance model online which is often critical to account for drastic appearance changes in tracking scenarios.  It is worth noting that CFNet \cite{CFNet} and DCFNet \cite{DCFNet} interpret the correlation filters as a differentiable layer in a Siamese tracking framework, thus achieving an end-to-end representation learning. The main drawback is their unsatisfying performance. Therefore, there still is performance improvement space for real-time deep trackers.

\begin{figure*}[thpb]
  \centering
  \includegraphics[width=0.8\linewidth]{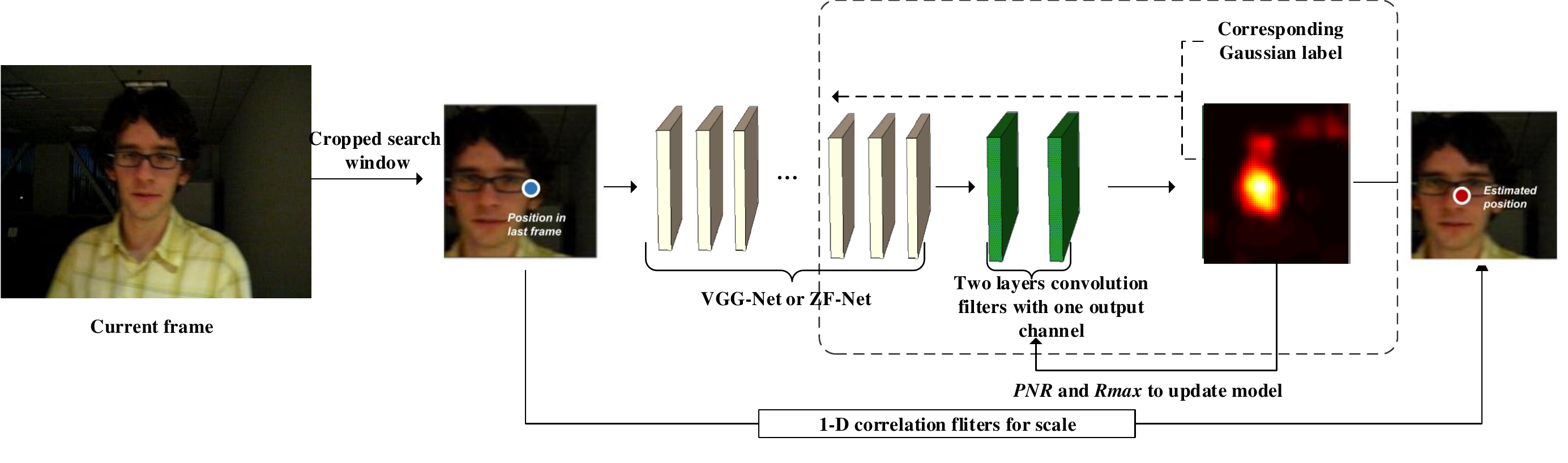}
  \caption{The overall UCT architecture. In each frame, patch is cropped around ground truth and resized into $224 \times 224$ with a padding of 1.8. The solid lines indicate online tracking process, while dashed box and dashed lines indicate off-line training and training on first frame.}
  \label{figure2}
\end{figure*}

%-------------------------------------------------------------------------
\subsection{On fully convolutional trackers}

Fully convolutional networks can efficiently learn to make dense predictions for visual tasks like semantic segmentation, detection as well as tracking. Paper \cite{semanticsegmentation} transforms fully connected layers into convolutional layers to output a heat map for semantic segmentation. The region proposal network (RPN) in Faster R-CNN \cite{FasterRCNN} is a fully convolutional network that simultaneously predicts object bounds and objectness scores at each position. DenseBox \cite{DenseBox} is an end-to-end FCN detection framework that directly predicts bounding boxes and object class confidences through whole image.
In tracking literatures, FCNT \cite{FCNT} proposes a two-stream fully convolutional network to capture both general and specific object information for visual tracking. However, its tracking components are still independent, so the performance may be impaired. In addition, the FCNT can only perform at 3 FPS on GPU because its layers switch mechanism and feature map selection method are time-consuming, which hinder it from real-time applications. Compared with FCNT, our UCT treats feature extractor and tracking process in a unified architecture and trains them end-to-end, resulting a more compact and much faster tracking approach.

\section{Unified Convolutional networks for tracking}

%\label{Unified}
In this section, the overall architecture of proposed UCT is introduced firstly. Afterwards, we detail the formulation of convolutional operation both in training and test stages. Lastly, training process of UCT is described.

\subsection{UCT Architecture}
The overall framework of our tracking approach is shown in Figure~\ref{figure2}, which consists of feature extractor and convolutions performing tracking process. The solid lines indicate online tracking process, while dashed box and dashed lines are included in off-line training and training on first frame. The search window of current frame is cropped and sent to unified convolutional networks. The estimated new target position is obtained by finding the maximum value of the response map. Another separate 1-dimensional convolutional branch is used to estimate target scale and model updating is performed if necessary, which enables our tracker can perform at real-time rate. Each feature channel in the extracted sample is always multiplied by a Hann window, as described in \cite{KCF}. The proposed framework has two advantages: First, we adopt two groups  convolutional filters to perform tracking process which is trained with features extractor. Compared to two-stage approaches adopted in DCF framework within CNN features \cite{HCF, HDT, DeepSRDCF}, our end-to-end training pipeline is generally preferable. The reason is that the parameters in all components can cooperate to achieve tracking objective. Second, during online tracking, the whole patch can be predicted using the foreground heat map by one-pass forward propagation. Redundant computation is saved. Whereas in \cite{MDNet} and \cite{DLT}, network has to be evaluated for $N$ times given $N$ samples cropped from the frame. The overlap between patches leads to a lot of redundant computation. Besides, we adopt efficient scale handling and model updating strategies to ensure real-time speed.

\subsection{Formulation}
In the UCT formulation, the aim is to learn a series of convolution filters $f$ from training samples ${(x_k, y_k)}_{k=1:t}$. Each sample is extracted using another CNN from an image region. Assuming sample has the spatial size $M \times N$, the output has the spatial size $m \times n$ ($m=M / stride_M, n=N / stride_N$). The desired output $y_k$ is a response map which includes a target score for each location in the sample $x_k$. The convolutional response of the filter on sample $x$ is given by
\begin{equation}
\label{eq1}
R(x) = \sum_{l=1}^dx^l*f^l
\end{equation}
where $x^l$ and $f^l$ is $l$-th channel of extracted CNN features and desired filters, respectively, $*$ denotes convolutional operation. The filter can be trained by minimizing $L_2$ loss which is obtained between the response $R(x_k)$ on sample $x_k$ and the corresponding Gaussian label $y_k$
\begin{equation}
\label{eq2}
L_2 = {||R(x_k) - y_k||}^2 + \lambda\sum_{l=1}^d{||f^l||}^2
\end{equation}
The second term in~(\ref{eq2}) is a regularization with a weight parameter $\lambda$.

In test stage, the trained filters are used to evaluate an image patch centered around the predicted target location. The evaluation is applied in a sliding-window manner, thus can be operated as convolution:
\begin{equation}
\label{eq3}
R(z) = \sum_{l=1}^dz^l*f^l
\end{equation}
where $z$ denote the feature map extracted from last target position including context.

It is noticed that the formulation in our framework is similar to DCF, which solve this ridge regression problem in frequency domain by circularly shifting the sample. Different from DCF, we adopt gradient descent to solve equation~(\ref{eq2}), resulting in convolution operations. Noting that the sample $x_k$ is also extracted by CNN, these convolution operations can be naturally unified in a fully convolutional network. Compared to DCF framework, our approach has three advantages: firstly, both feature extraction and tracking convolutions can be pre-trained simultaneously, while DCF based trackers can only tune the hyper-parameters heuristically. Secondly, model updating can be performed by stochastic gradient descent (SGD), which maintains the long-term memory of target appearance. Lastly, our framework is much faster than DCF framework within CNN features.
\subsection{Training}
Since the objective function defined in equation~(\ref{eq2}) is convex, it is possible to obtain the approximate global optima via gradient descent with an appropriate learning rate in limited steps. We divide the training process into two periods: off-line training that can encode the prior tracking knowledge, and the training on first frame to adapt to specific target.

In off-line training, the goal is to minimize the loss function in equation~(\ref{eq2}). In tracking, the target position in last frame is always not centered in current cropped patch. So for each image, the train patch centered at the given object is cropped with jittering. The jittering consists of translation and scale jittering, which approximates the variation in adjacent frames when tracking.   Above cropped patch also includes background information as context. In training, the final response map is obtained by last convolution layer within one channel. The label is generated using a Gaussian function with variances proportional to the width and height of object. Then the $L_2$ loss can be generated and the gradient descent can be performed to minimize equation~(\ref{eq2}). In this stage, the overall network consists of a pre-trained network with ImageNet (VGG-Net in UCT and ZF-Net in UCT-lite) and following convolutional filters. Last part of VGG-Net or ZF-Net is trained to encode the prior tracking knowledge with following convolutional filters, making the extracted feature more suitable for tracking.

\begin{figure}[thpb]
  \centering
  \includegraphics[width=1\linewidth]{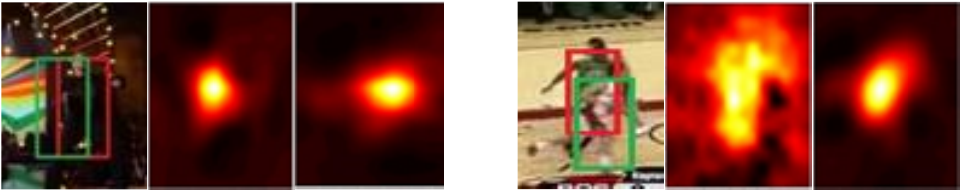}
  \caption{From left to right: images, response maps without off-line training and response maps with off-line training. Green and red boxes in images indicates tracking results without and with off-line training, respectively.}
  \label{figure3}
\end{figure}

The goal of training on first frame is to adapt to a specific target. The network architecture follows that in off-line training, while later convolutional filters are randomly initialized by zero-mean Gaussian distribution. Only these randomly initialized layers are trained using SGD in first frame. Offline training encodes prior tracking knowledge and constitutes a tailored feature extractor. We perform online tracking with and without off-line training to illustrate this effect. In Figure~\ref{figure3}, we show tracking results and corresponding response maps without or with offline training. In left part of Figure~\ref{figure3}, the target singer is moving to right, the response map with off-line training effectively reflects this translation changes while response map without off-line training is not capable of doing this. So the tracker without off-line training misses this critical frame. In right part of Figure~\ref{figure3}, the target player is occluded by another player, the response map without off-line training becomes fluctuated and tracking result is effected by distractor, while response map with off-line training still keeps discriminative. The results are somewhat unsurprising, since CNN features trained on ImageNet classification data are expected to have greater invariance to position and same class. In contrast, we can obtain more suitable feature tracking by end-to-end off-line training.

\section{Online tracking}

After off-line training and training on first frame, the learned network is used to perform online tracking by equation~(\ref{eq3}). The estimate of the current target state is obtained by finding the maximum response score. Since we use a fully convolutional network architecture to perform tracking, the whole patch can be predicted using the foreground heat map by one-pass forward propagation. Redundant computation was saved. Whereas in \cite{MDNet} and \cite{DLT}, network has to be evaluated for $N$ times given $N$ samples cropped from the frame. The overlap between patches leads to a lot of redundant computation.

\subsection{Model update}
Most of tracking approaches update their models in each frame or at a fixed interval \cite{KCF, CSK, HCF, CCOT}. However, this strategy may introduce false background information when the tracking is inaccurate, target is occluded or out of view.  In the proposed method, model update is decided by evaluating the tracking results. Specifically, we consider the maximum value in the response map and the distribution of other response value simultaneously.

\begin{figure}[thpb]
  \centering
  \includegraphics[width=0.8\linewidth]{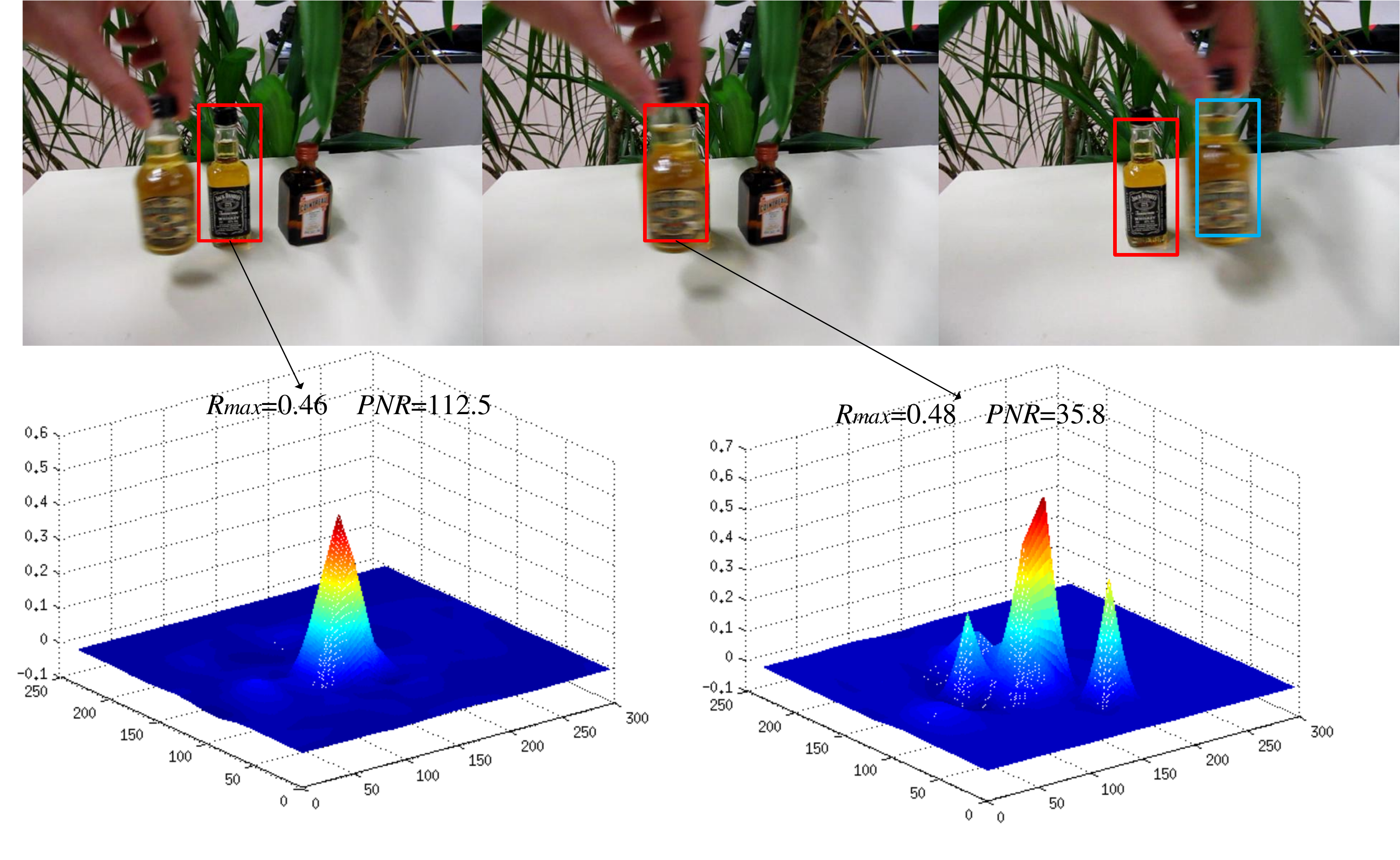}
  \caption{Updating results of UCT and UCT$\_$No$\_$\emph{PNR} (UCT without \emph{PNR} criterion). The first row shows frames that the target is occluded by distractor. The second row is corresponding response maps. $R_{max}$ still keeps large in occlusion while \emph{PNR} significantly decreases. So the unwanted updating is avoided by considering \emph{PNR} constraint simultaneously. The red and blue boxes in last image are tracking results of UCT and UCT$\_$No$\_$\emph{PNR}, respectively.}
  \label{figure4}
\end{figure}

Ideal response map should have only one peak value in actual target position and the other values are small.  On the contrary, the response will fluctuate intensely and include more peak values as shown in Figure~\ref{figure4}. We introduce a novel criterion called peak-versus-noise ratio (\emph{PNR}) to reveal the distribution of response map. The \emph{PNR} is defined as
{\setlength\abovedisplayskip{5pt}
\setlength\belowdisplayskip{5pt}
\begin{equation}
\label{eq4}
\emph{PNR} = \frac{R_{max}-R_{min}}{mean(R{\backslash}R_{max})}
\end{equation}}where
{\setlength\abovedisplayskip{1pt}
\setlength\belowdisplayskip{1pt}
\begin{equation}
R_{max} = \max{R(z)}
\end{equation}}and $R_{min}$ is corresponding minimum value of response map. Denominator in equation~(\ref{eq4}) represents mean value of response map except maximum value and it is used to measure the noise approximately. The \emph{PNR} becomes larger when response map has fewer noise and sharper peak. Otherwise, the \emph{PNR} will fall into a smaller value. We save the \emph{PNR} and $R_{max}$ and calculate their historical average values as threshold:
{\setlength\abovedisplayskip{8pt}
\setlength\belowdisplayskip{8pt}
\begin{equation}
\label{eq6}
\left\{
     \begin{array}{rl}
     \emph{PNR}_{threshold} =& \frac{\sum_{t=1}^T\emph{PNR}_t}{T}  \\
     R_{threshold} =& \frac{\sum_{t=1}^T{R}_{max}^t}{T}
     \end{array}
\right.
\end{equation}}Model update is performed only when \emph{PNR} and $R_{max}$ are larger than corresponding threshold at the same time. The updating is one step SGD with smaller learning rate compared with that in the first frame. Figure~\ref{figure4} illustrates the necessity of proposed \emph{PNR} criterion by showing tracking results under occlusions. As shown in Figure~\ref{figure4}, updating is still performed if only $R_{max}$ is adopted. Introduced noise will result in inaccurate tracking results even failures. The \emph{PNR} value significantly decreases in these unreliable frames thus avoids unwanted updating.

\subsection{Scale estimation}

A conventional approach of incorporating scale estimation is to evaluate the appearance model at multiple resolutions by performing an exhaustive scale search \cite{SAMF}. However, this search strategy is computationally demanding and not suitable for real-time tracking. Inspired by \cite{fDSST}, we introduce a 1-dimensional convolutional filters branch to estimate the target size as shown in Figure~\ref{figure2}.  This scale filter is applied at an image location to compute response scores in the scale dimension, whose maximum value can be used to estimate the target scale. Such learning separate convolutional filters to explicitly handle the scale changes is more efficient for real-time tracking.

In training and updating of scale convolutional filters, the sample $x$ is extracted from variable patch sizes centered around the target:
{\setlength\abovedisplayskip{1pt}
\setlength\belowdisplayskip{1pt}
\begin{equation}
\label{eq7}
size(P^n) = a^nW \times a^nH~~~n \in \{-\lfloor\frac{S-1}{2}\rfloor,...,\lfloor\frac{S-1}{2}\rfloor\}
\end{equation}}where $S$ is the size of scale convolutional filters, $W$ and $H$ are the current target size, $a$ is the scale factor. In scale estimation test, the sample is extracted using the same way after translation filters are performed. Then the scale changes compared to previous frame can be obtained by maximizing the response score. Note that the scale estimation is performed only when model updating conditions are satisfied.

The overall tracking algorithm is summarized in Algorithm 1.

\begin{algorithm}[htb]
\caption{ Unified Convolutional Networks for Tracking}
\label{algorithm1}
\begin{algorithmic}[1] %
\REQUIRE ~~\\ %
Initial target position, \bm{$P_0$};\\
Off-line learned filters \bm{$w$};\\
\ENSURE Target position \bm{$P_t$} during sequences\\ %
\REPEAT
\IF{first frame}
\STATE fine-tune the network using ground truth by (2)
\ELSE
\STATE Crop out new patch \bm{$z$} centered at previous results
\STATE Calculate the response map $R(z)$ using (3)
\STATE Calculate the scale changes compared to previous frame by (7)
\STATE Obtain the bounding box \bm{$P_t$} in current frame according to $R_{max}$ and scale changes
\STATE Calculate the $\emph{PNR}_{threshold}$ and ${R}_{threshold}$ using (4) and (6)
\IF{ two criterions are satisfied}
\STATE update the model using (2)
\ENDIF
\ENDIF
\UNTIL{End of video sequences;}
\end{algorithmic}
\end{algorithm}

\section{Experiments}
Experiments are performed on four challenging tracking datasets: OTB2013~\cite{OTB2013}, OTB2015~\cite{OTB2015}, VOT2015~\cite{VOT2015} and VOT2016~\cite{VOT2016}. Basic information about four dataset is summarized in Table~\ref{table_dataset}. All the tracking results use the reported results to ensure a fair comparison.

\begin{table}[t]
  \centering
  \caption{Basic information about four dataset in experiments. IV: Illumination Variation. OPR: Out-of-Plane Rotation. SV: Scale Variation. OCC: Occlusion. DEF: Deformation. MB: Motion Blur. FM: Fast Motion. IPR: In-Plane Rotation. OV: Out-of-View. BC: Background Clutters. LR: Low Resolution. CM: Camera Motion. MC: Motion Change.  }
 \begin{tabular}{c|c|c}
    \hline
     Dataset   &  Frame number  &   Main challenges \\
    %\midrule
    \hline
    OTB2013   & 29491 &  \shortstack{IV,OPR,SV,OCC,DEF\\MB,FM,IPR,OV,BC,LR}   \\\hline
    OTB2015   & 59040 &  \shortstack{IV,OPR,SV,OCC,DEF\\MB,FM,IPR,OV,BC,LR}    \\\hline
    VOT2015   & 21871 &  CM,IV,OCC,SV,MC   \\\hline
    VOT2016     & 21871 &  CM,IV,OCC,SV,MC \\
    %\bottomrule
    \hline

  \end{tabular}
   \label{table_dataset}
\end{table}

\subsection{Implementation details}

We adopt VGG-Net \cite{VGG} in standard UCT and ZF-Net \cite{ZFNet} in UCT-lite as feature extractor, respectively. In off-line training, last six layers of VGG-Net and last three layers of ZF-Net are fine-tuned. The kernel size of two convolutional layers for tracking is $7\times7$ and the activation function is Sigmoid. Our training data comes from VID \cite{ILSVRC15}, containing the training and validation set. In each frame, patch is cropped around ground truth and resized into $224\times224$ with a padding of 1.8. The translation and scale jittering are 0.05 and 0.02 proportional to the size of images, respectively.  We apply SGD with momentum of 0.9 to train the network and set the weight decay $\lambda$ to 0.005. The model is trained for 30 epochs with a learning rate of $10^{-5}$. In online training on first frame, SGD is performed 50 steps with a learning rate of $5\times10^{-7}$ and $\lambda$ is set to 0.01. In online tracking, the model updating is performed by one step SGD with a learning rate of $10^{-7}$. $S$ and $a$ in equation~(\ref{eq7}) is set to 33 and 1.02, respectively.

The proposed UCT is implemented using Caffe \cite{Caffe} with Matlab wrapper on a PC with an Intel i7 6700 CPU, 48 GB RAM, Nvidia GTX TITAN X GPU. The code and results will be made publicly available.

\subsection{Results on OTB2013}
\label{resultsonOTB2013}
OTB2013 \cite{OTB2013} contains 50 fully annotated sequences that are collected from commonly used tracking sequences. The evaluation is based on two metrics: precision plot and success plot. The precision plot shows the percentage of frames that the tracking results are within certain distance determined by given threshold to the ground truth. The value when threshold is 20 pixels is always taken as the representative precision score. The success plot shows the ratios of successful frames when the threshold varies from 0 to 1, where a successful frame means its overlap is larger than this given threshold. The area under curve (AUC) of each success plot is used to rank the tracking algorithms.

In this experiment, ablation analyses are performed to illustrate the effectiveness of proposed component at first. Then we compare our method against the recent real-time (and near real-time) trackers presented at top conferences and journals, including PTAV \cite{PTAV}, BACF \cite{BACF}, CFNet \cite{CFNet}, CACF \cite{CACF}, CSR-DCF \cite{CSRDCF}, fDSST \cite{fDSST}, SiamFC \cite{SiamFC}, Staple \cite{Staple}, SCT \cite{SCT}, HDT \cite{HDT}, HCF \cite{HCF}, LCT \cite{LCT}, KCF \cite{KCF}. The one-pass evaluation (OPE) is employed to compare these trackers.

\begin{figure}[!tp]
 \centering
\begin{minipage}[c]{8cm}
\includegraphics[width=8cm]{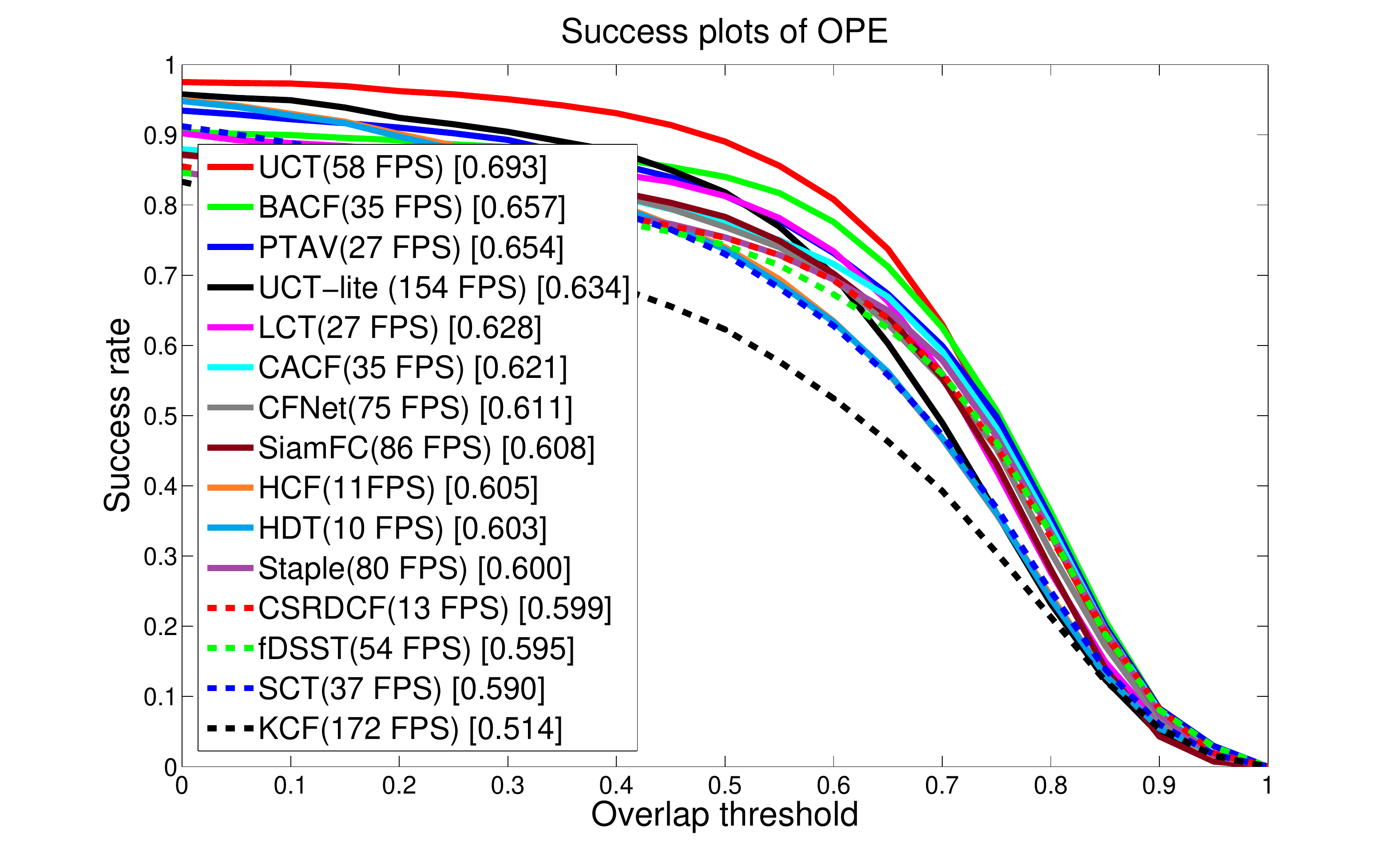}
\end{minipage}%

\begin{minipage}[c]{8cm}
\includegraphics[width=8cm]{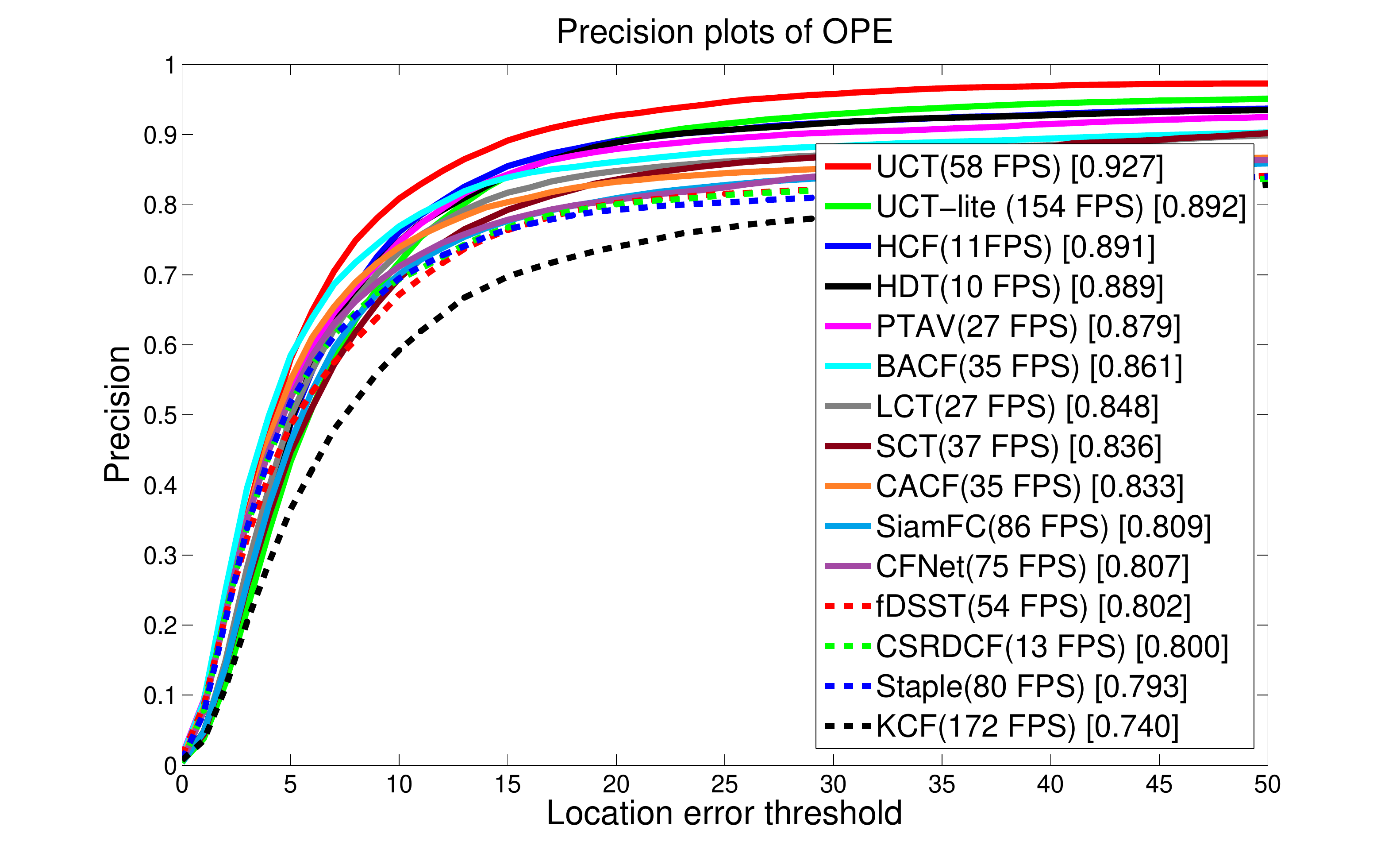}
\end{minipage}%
 \caption{Precision and success plots on OTB2013. The numbers in the legend indicate the representative precisions at 20 pixels for precision plots, and the area-under-curve scores for success plots. Best viewed on color display.}
 \label{OTB2013_OPE}
\end{figure}

\begin{figure*}[thpb]
 \centering
\begin{minipage}[c]{4.7cm}
\includegraphics[width=5cm]{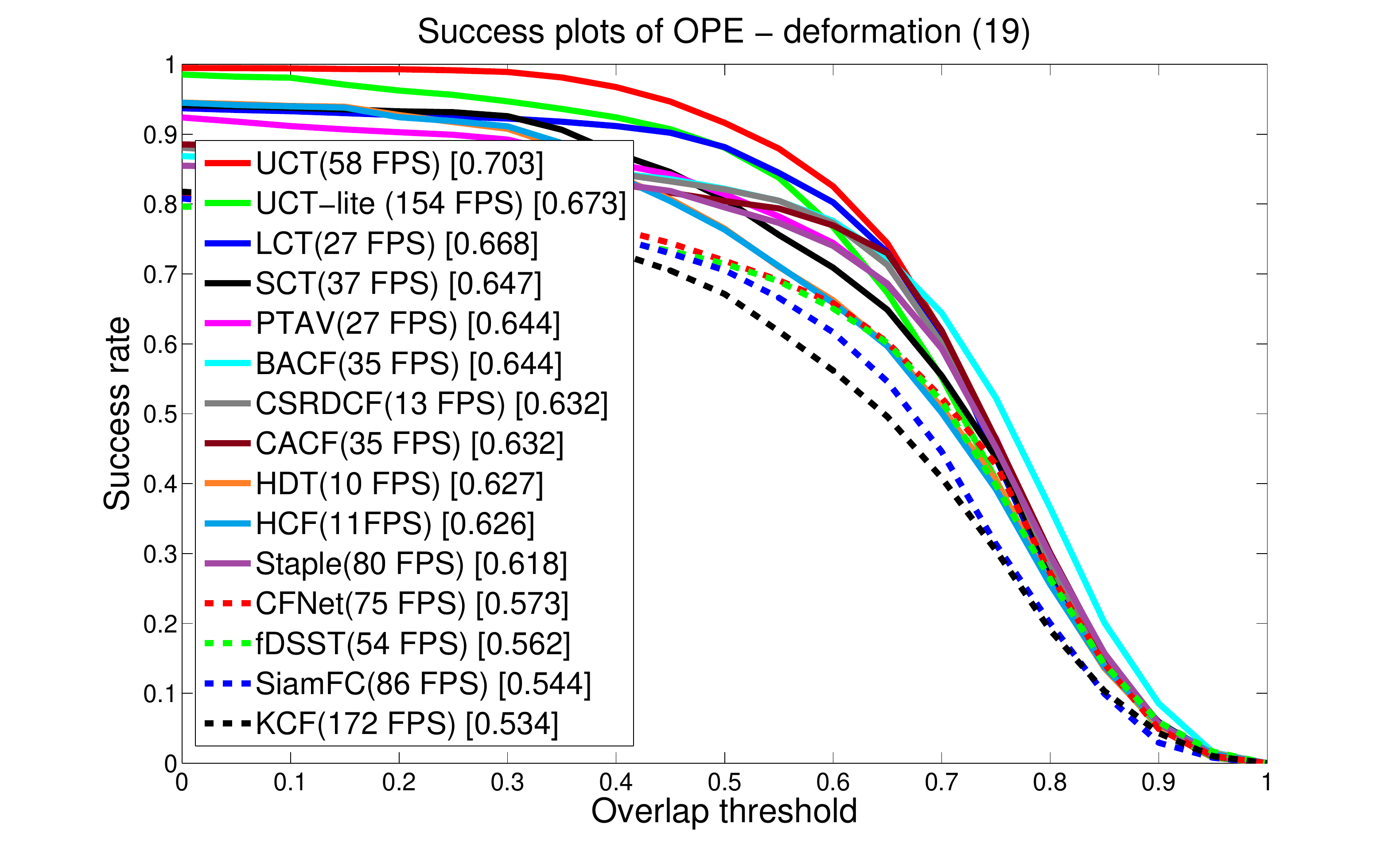}
\end{minipage}%
\begin{minipage}[c]{4.7cm}
\includegraphics[width=5cm]{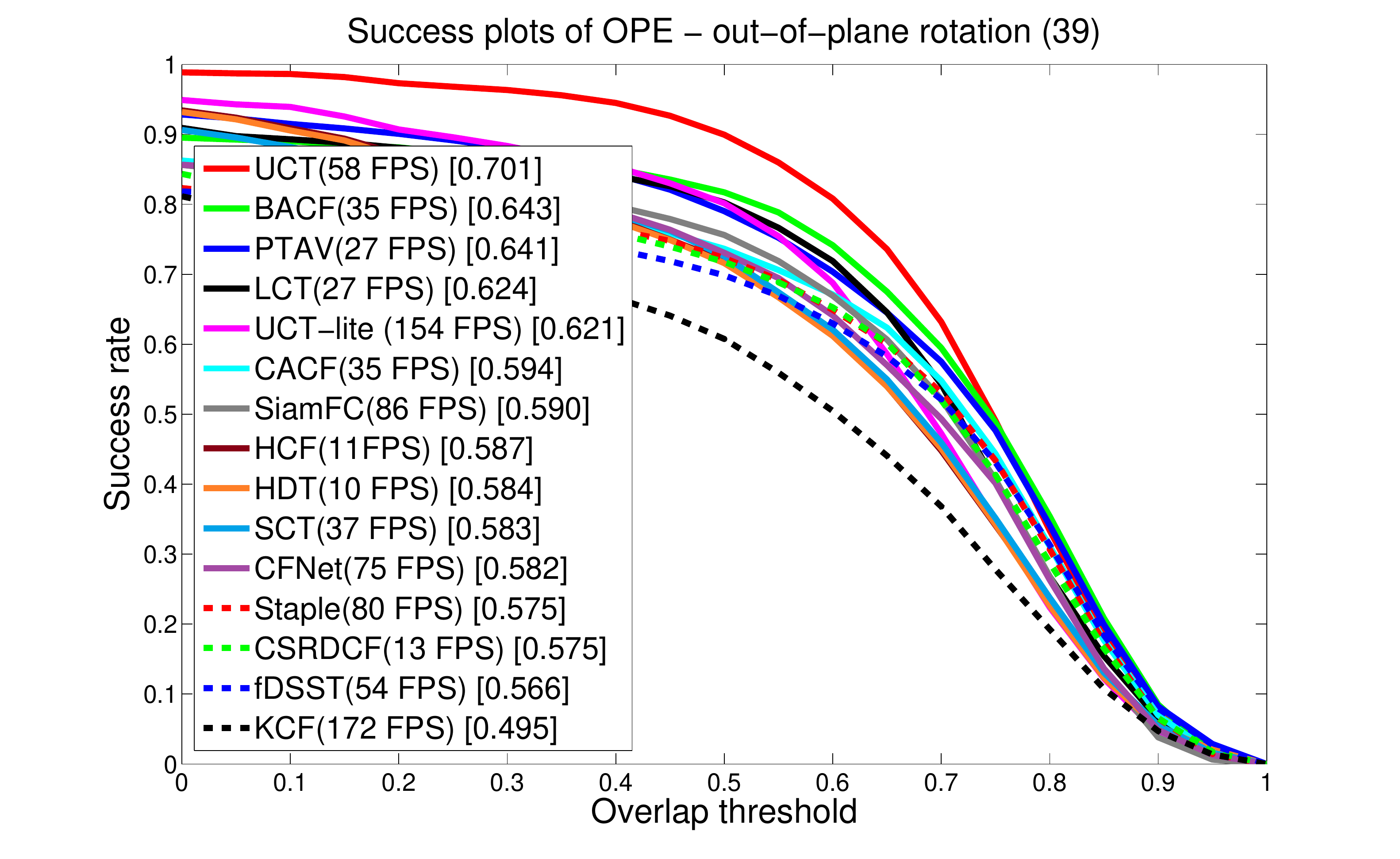}
\end{minipage}%
\begin{minipage}[c]{4.7cm}
\includegraphics[width=5cm]{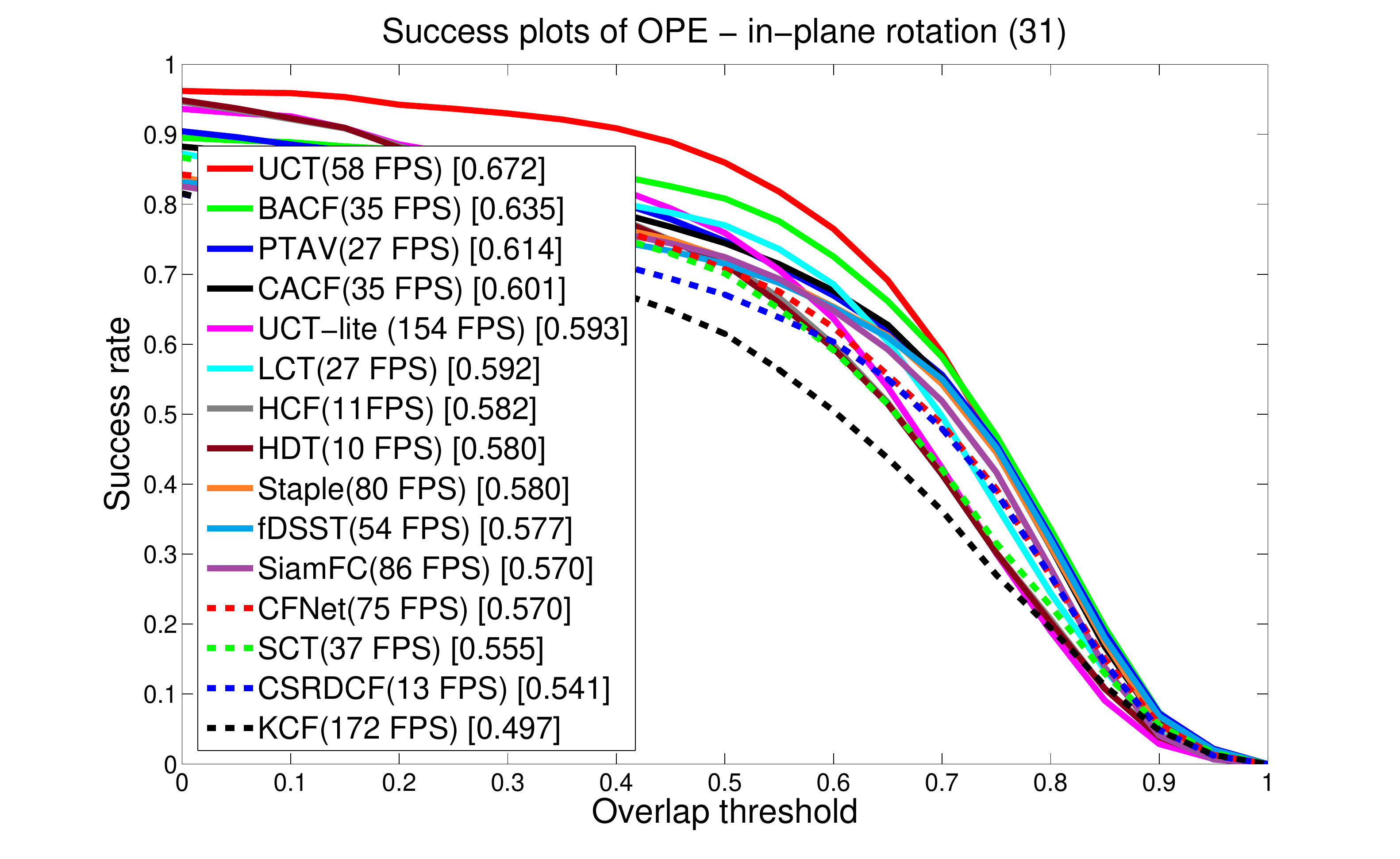}
\end{minipage}%
\begin{minipage}[c]{4.7cm}
\includegraphics[width=5cm]{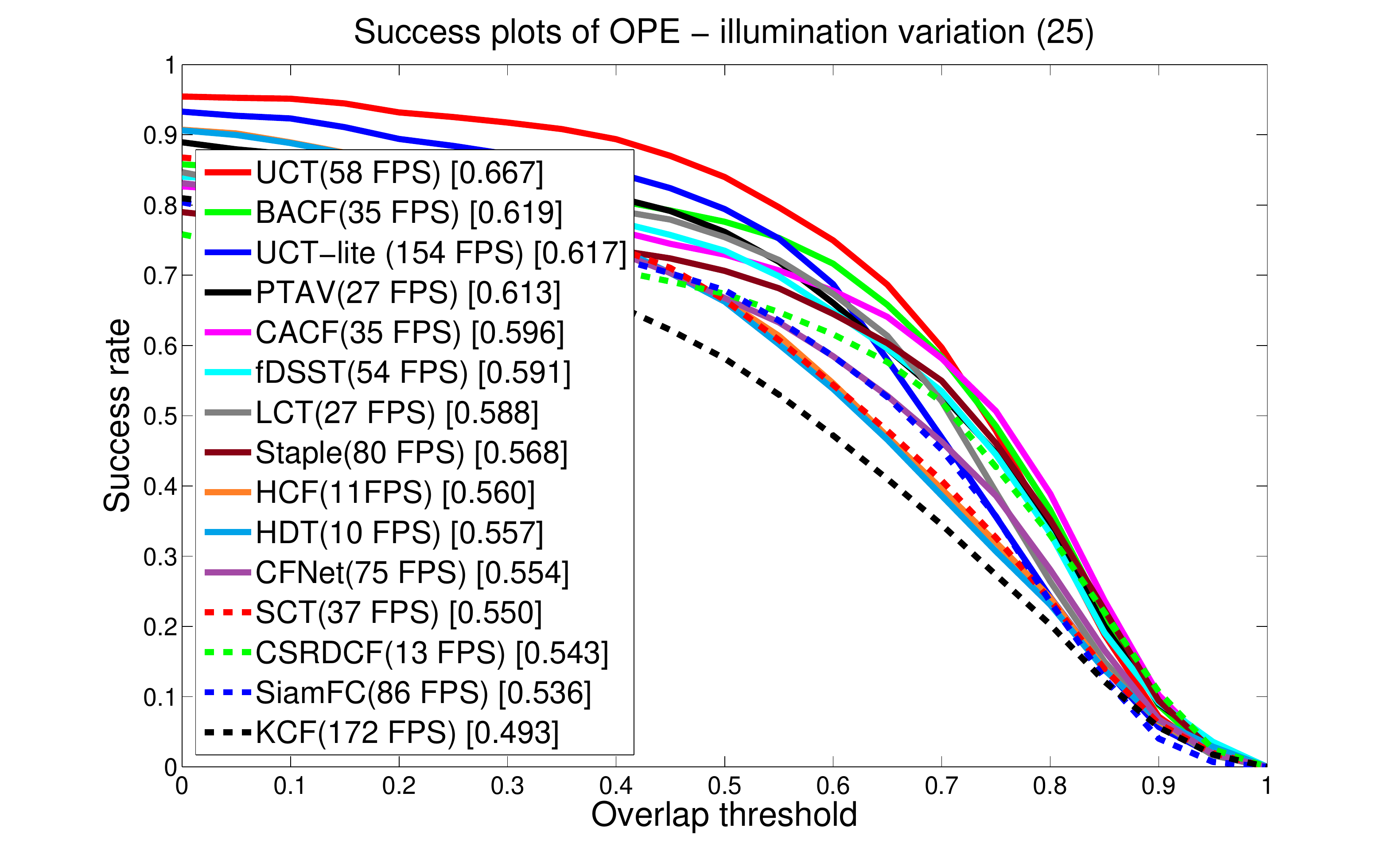}
\end{minipage}%

\begin{minipage}[c]{4.7cm}
\includegraphics[width=5cm]{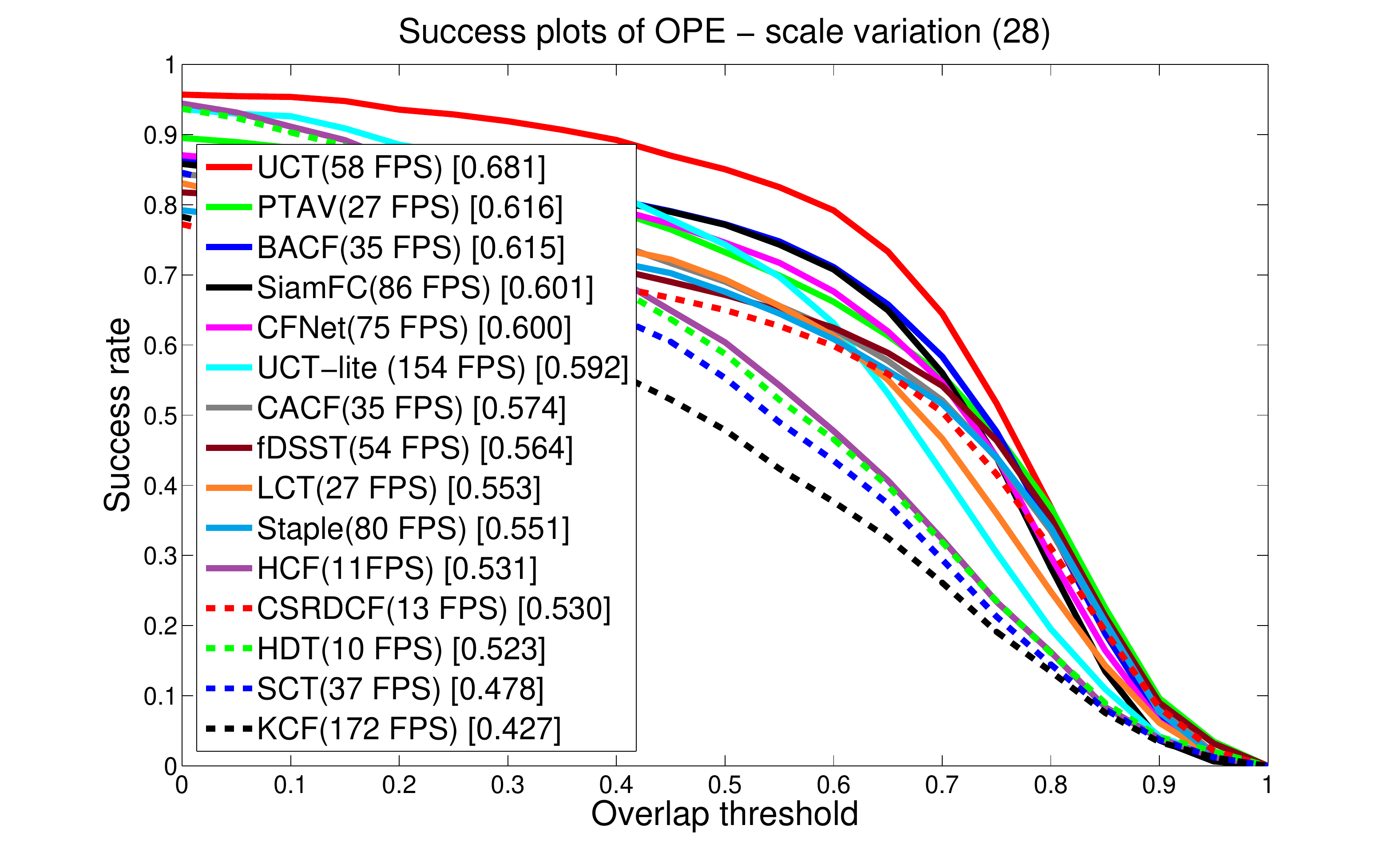}
\end{minipage}%
\begin{minipage}[c]{4.7cm}
\includegraphics[width=5cm]{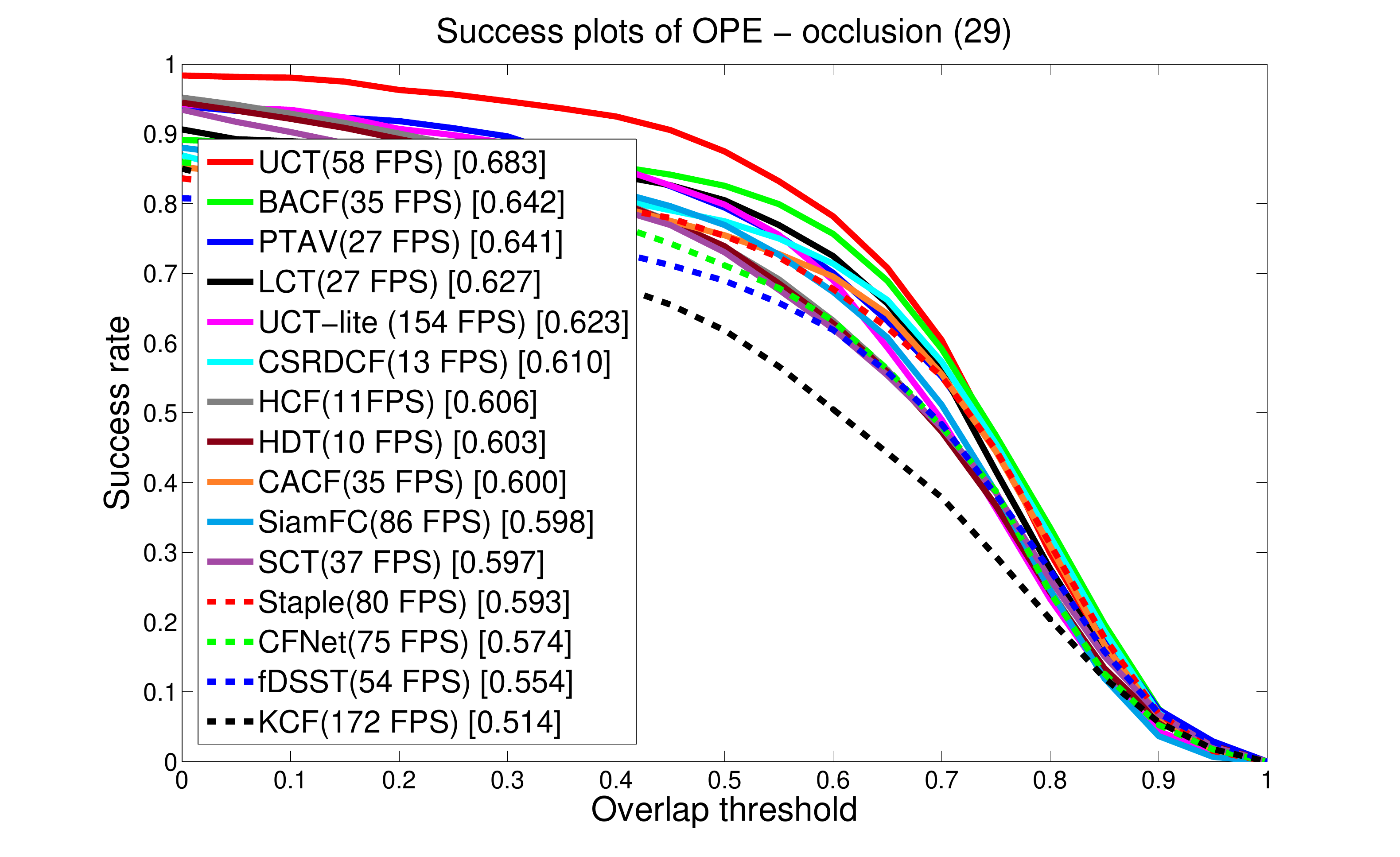}
\end{minipage}%
\begin{minipage}[c]{4.7cm}
\includegraphics[width=5cm]{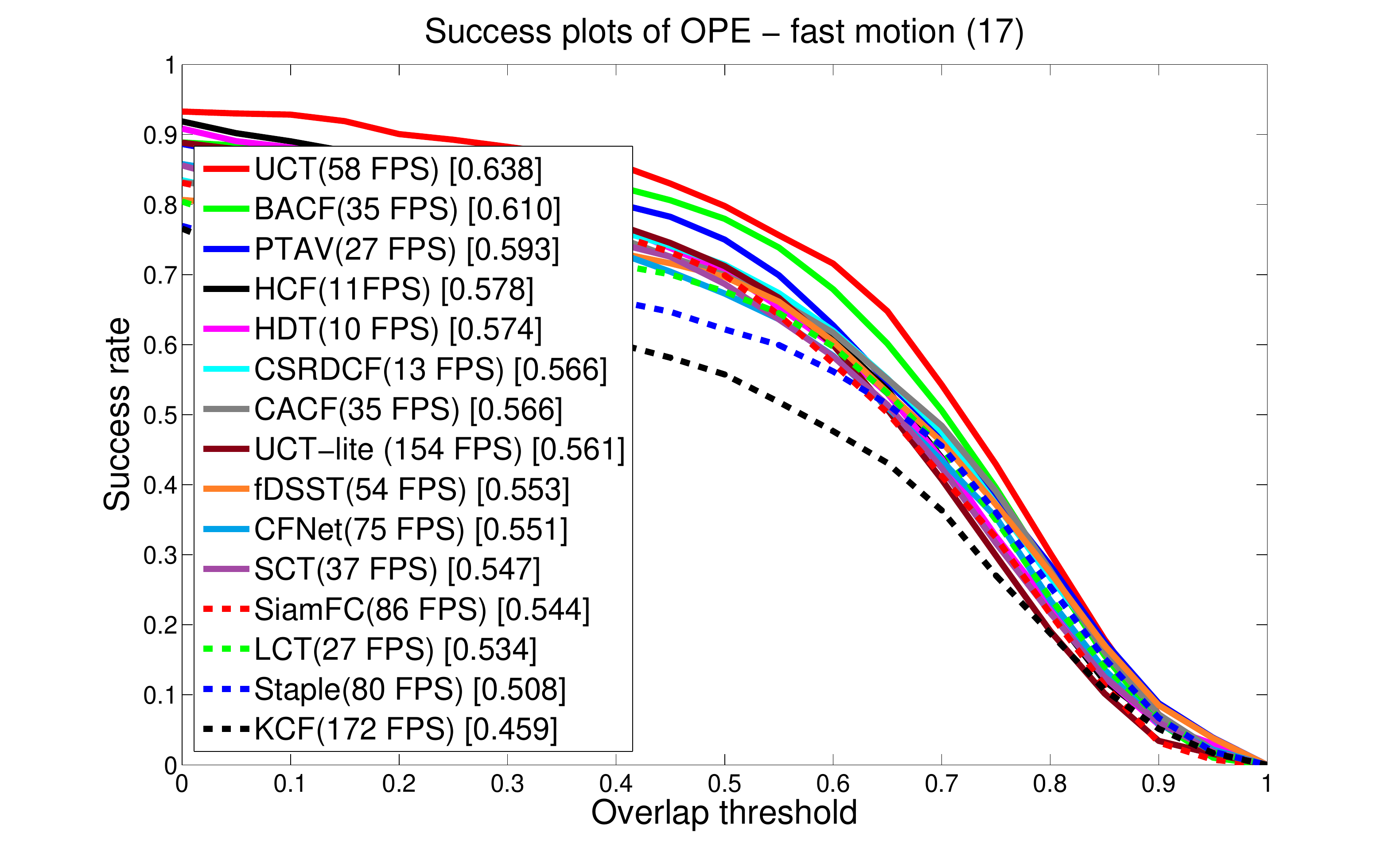}
\end{minipage}%
\begin{minipage}[c]{4.7cm}
\includegraphics[width=5cm]{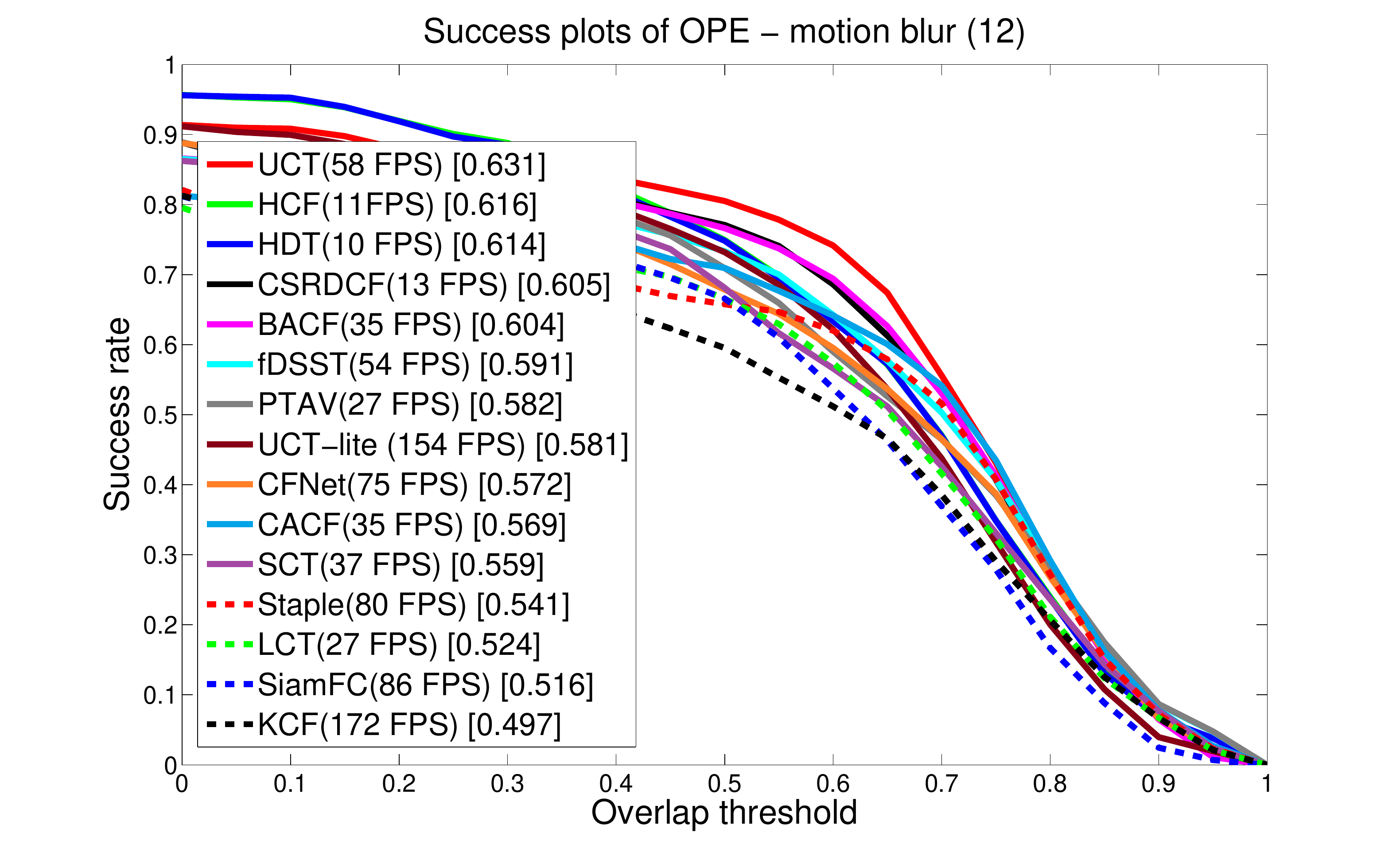}
\end{minipage}%
 \caption{Attributes performance on OTB2013. Best viewed on color display.}
  \label{OTB2013_OPE_attributes}
\end{figure*}

\subsubsection{\textbf{Ablation analyses}}
To verify the contribution of each component in our algorithm, we implement and evaluate four variations of our approach: firstly, the effectiveness of our off-line training is tested by comparison without this procedure (\emph{UCT$\_$No$\_$Off-line}), where the network is only trained within the first frame of a specific sequence. Secondly, the tracking algorithm that updates model without \emph{PNR} constraint (\emph{UCT$\_$No$\_$PNR}, only depends on $R_{max}$) is compared with the proposed efficient updating method. Last two additional versions are UCT within multi-resolutions scale (\emph{UCT$\_$MulRes$\_$Scale}) and without scale handling (\emph{UCT$\_$No$\_$Scale}).

As shown in Table~\ref{table2}, the performances of all the variations are not as good as our full algorithm (UCT) and each component in our tracking algorithm is helpful to improve performance. Specifically, off-line training encodes prior tracking knowledge and constitutes a tailored feature extractor, so the UCT outperforms \emph{UCT$\_$No$\_$Off-line} with a large margin. Proposed \emph{PNR} constraint for model update improves performance as well as speed, since it avoids updating in unreliable frames. Although exhaustive scale method in multiple resolutions improves the performance of tracker, it brings higher computational cost. By contrast, learning separate filters for scale in our approach gets a better performance while being computationally efficient.

\begin{table}[t]
  \centering
  \caption{ Performance on OTB2013 of UCT and its variations}
 \begin{tabular}{cccc}
    \hline
    \bf Approaches     & \bf AUC     & \bf Precision20 & \bf Speed (FPS) \\
    %\midrule
    \hline
    UCT$\_$no$\_$off-line & 0.629  & 0.879 &  58     \\
    UCT$\_$no$\_$\emph{PNR} & 0.676  & 0.912 &  37      \\
    UCT$\_$no$\_$scale    & 0.654  & 0.901 &  69     \\
    UCT$\_$mulRes$\_$scale& 0.664  & 0.909 &  27     \\
     \hline
    UCT                   & \textbf{0.693}  & \textbf{0.927} &  58     \\
    UCT-lite               & 0.634  & 0.892 &  \textbf{154}     \\
    %\bottomrule
    \hline

  \end{tabular}

   \label{table2}
\end{table}

\subsubsection{\textbf{Comparison with state-of-the-art trackers}}
We compare our method against the state-of-the-art trackers as listed earlier. Figure~\ref{OTB2013_OPE} illustrates the precision and success plots based on center location error and bounding box overlap ratio, respectively. It clearly illustrates that our algorithm, denoted by UCT, outperforms the state-of-the-art trackers significantly in both measures. In success plot, our approach obtains an AUC score of 0.693, significantly outperforms BACF and SiamFC by 3.6\% and 8.5\%, respectively.  In precision plot, our approach obtains a score of 0.927, outperforms BACF and SiamFC by 6.6\% and 11.8\%, respectively. We summarise the model update frequency, scale estimation, performance and speed of compared trackers in Table~\ref{table_all_otb}.

Besides standard UCT, we also implement a lite version of UCT (UCT-lite) which adopts ZF-Net \cite{ZFNet} and ignores scale changes. As shown in Figure~\ref{OTB2013_OPE}, the UCT-lite obtains a success score of 0.634 and precision score of 0.892, while operates at 154 FPS. Our UCT-lite approach is much faster than recent real-time trackers such as CFNet, SiamFC and Staple, while significantly outperforms them in performance.

Furthermore, the results on various challenge attributes in OTB2013 are reported for detailed performance analysis. These challenges include scale variation, fast motion, background clutter, deformation, occlusion, etc. Figure~\ref{OTB2013_OPE_attributes} demonstrates that our tracker effectively handles these challenging situations while other trackers obtain lower scores. Qualitative comparisons of our approach with four state-of-the-art trackers in the changing scenario are shown in Figure~\ref{figure13}.
The top performance can be attributed to that our methods encodes prior tracking knowledge by off-line training and extracted features is more suitable for following tracking convolution operations. For example, our tracker ranks top in out-of-plane rotation and in-plane rotation challenges due to the end-to-end training.
By contrast, the CNN features in other trackers are always pre-trained in different tasks and are independently with the tracking process, thus the achieved tracking performance may not be optimal. Furthermore, efficient updating and scale handling strategies ensure robustness and speed of the tracker.

\subsection{Results on OTB2015}
OTB2015 \cite{OTB2015} is the extension of OTB2013 and contains 100 video sequences. Some new sequences are more difficult to track. In this experiment, we compare our method against recent real-time trackers, including PTAV \cite{PTAV}, BACF \cite{BACF}, CFNet \cite{CFNet}, fDSST \cite{fDSST}, SiamFC \cite{SiamFC}, Staple \cite{Staple}, HDT \cite{HDT}, HCF \cite{HCF}, LCT \cite{LCT}, KCF \cite{KCF}. The one-pass evaluation (OPE) is employed to compare these trackers.

\begin{figure}[!tp]
 \centering
\begin{minipage}[c]{7.5cm}
\includegraphics[width=7.5cm]{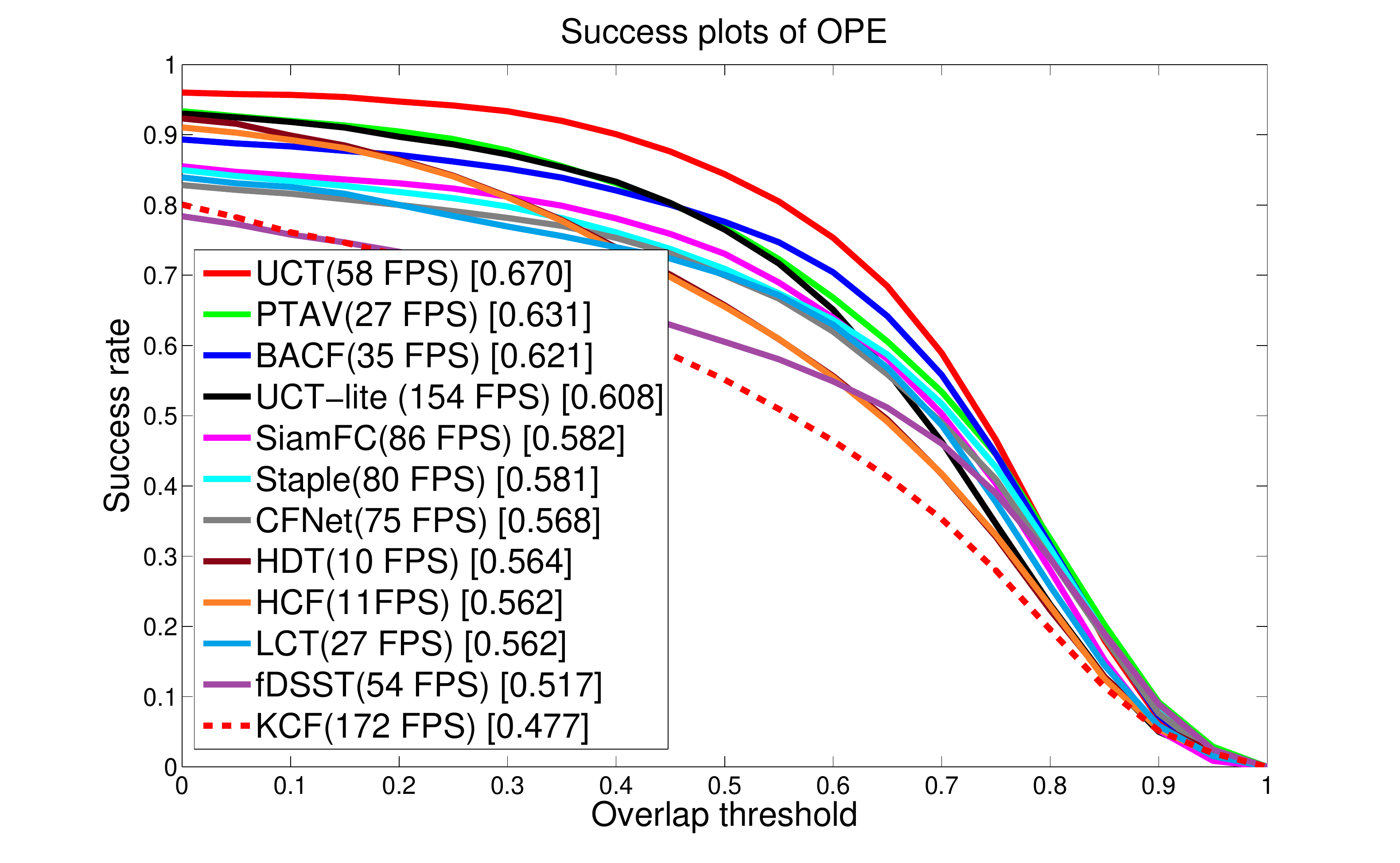}
\end{minipage}%

\begin{minipage}[c]{7.5cm}
\includegraphics[width=7.5cm]{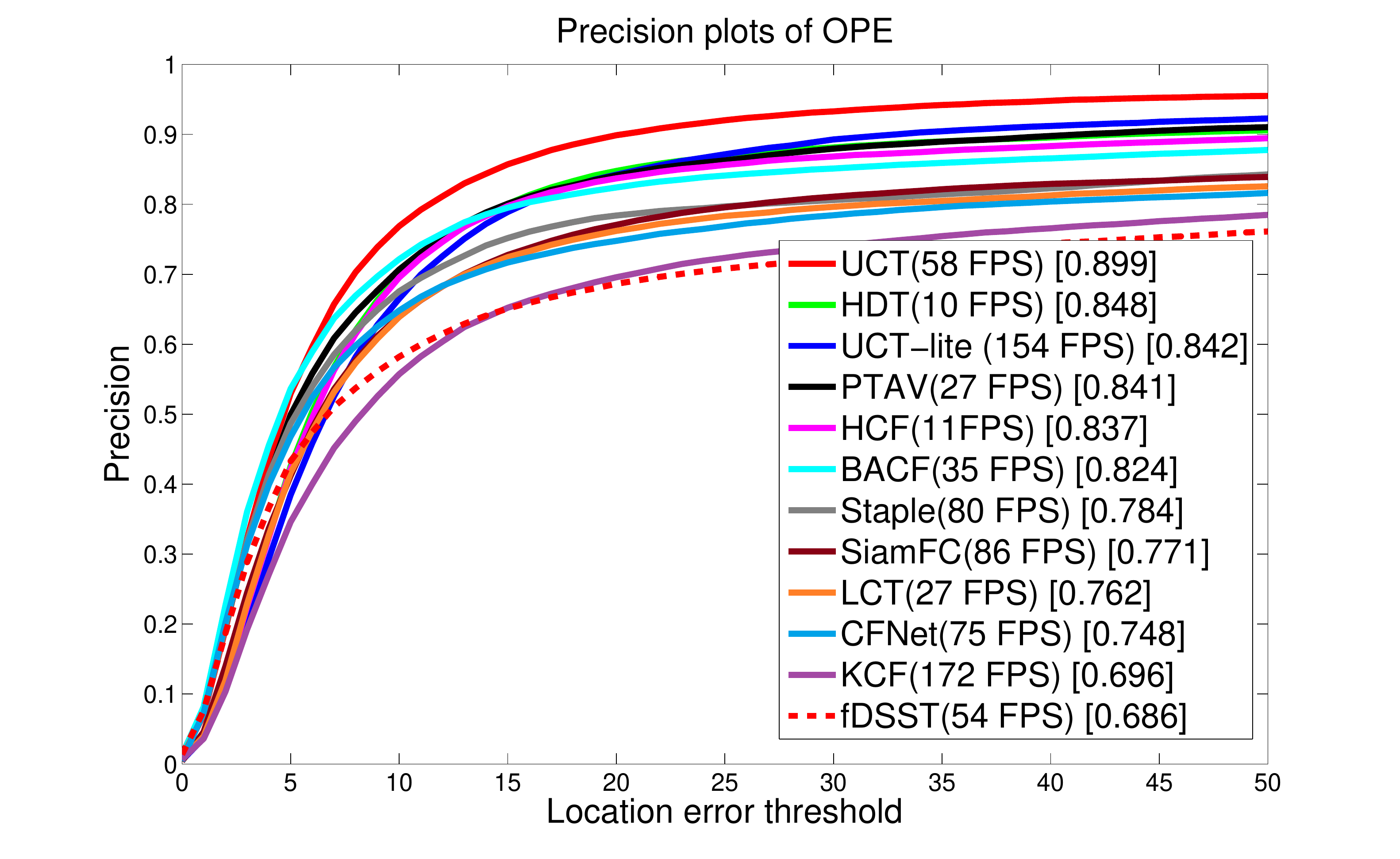}
\end{minipage}%
 \caption{Precision and success plots on OTB2015. The numbers in the legend indicate the representative precisions at 20 pixels for precision plots, and the area-under-curve scores for success plots. Best viewed on color display.}
 \label{OTB2015_OPE}
\end{figure}

Figure~\ref{OTB2015_OPE} illustrates the precision and success plots of the compared trackers, respectively. The proposed UCT approach outperforms all the other trackers in terms of both precision score and success score. Specifically, our method achieves a success score of 0.670, which outperforms the PTAV (0.631) and BACF (0.621) methods with a large margin. Since the proposed tracker adopts a unified convolutional architecture and efficient online tracking strategies, it achieves superior tracking performance and real-time speed. It is worth mentioning that our UCT-lite also provides significantly better performance and faster speed compared with SiamFC, CFNet and Staple. Model update frequency, scale estimation, performance and speed of compared trackers are summarised in Table~\ref{table_all_otb}.

\begin{figure*}[thpb]
 \centering
\begin{minipage}[c]{4.7cm}
\includegraphics[width=5cm]{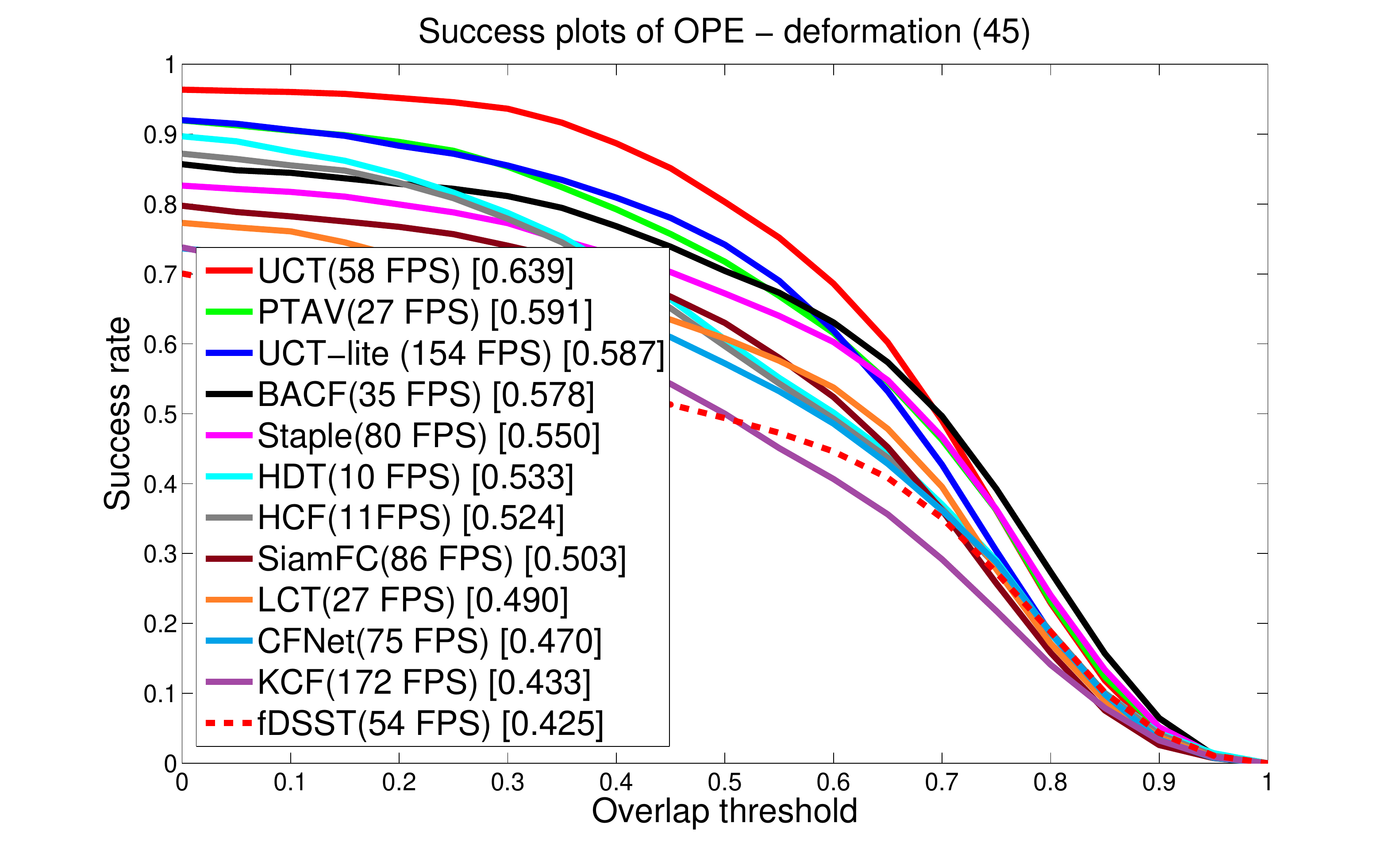}
\end{minipage}%
\begin{minipage}[c]{4.7cm}
\includegraphics[width=5cm]{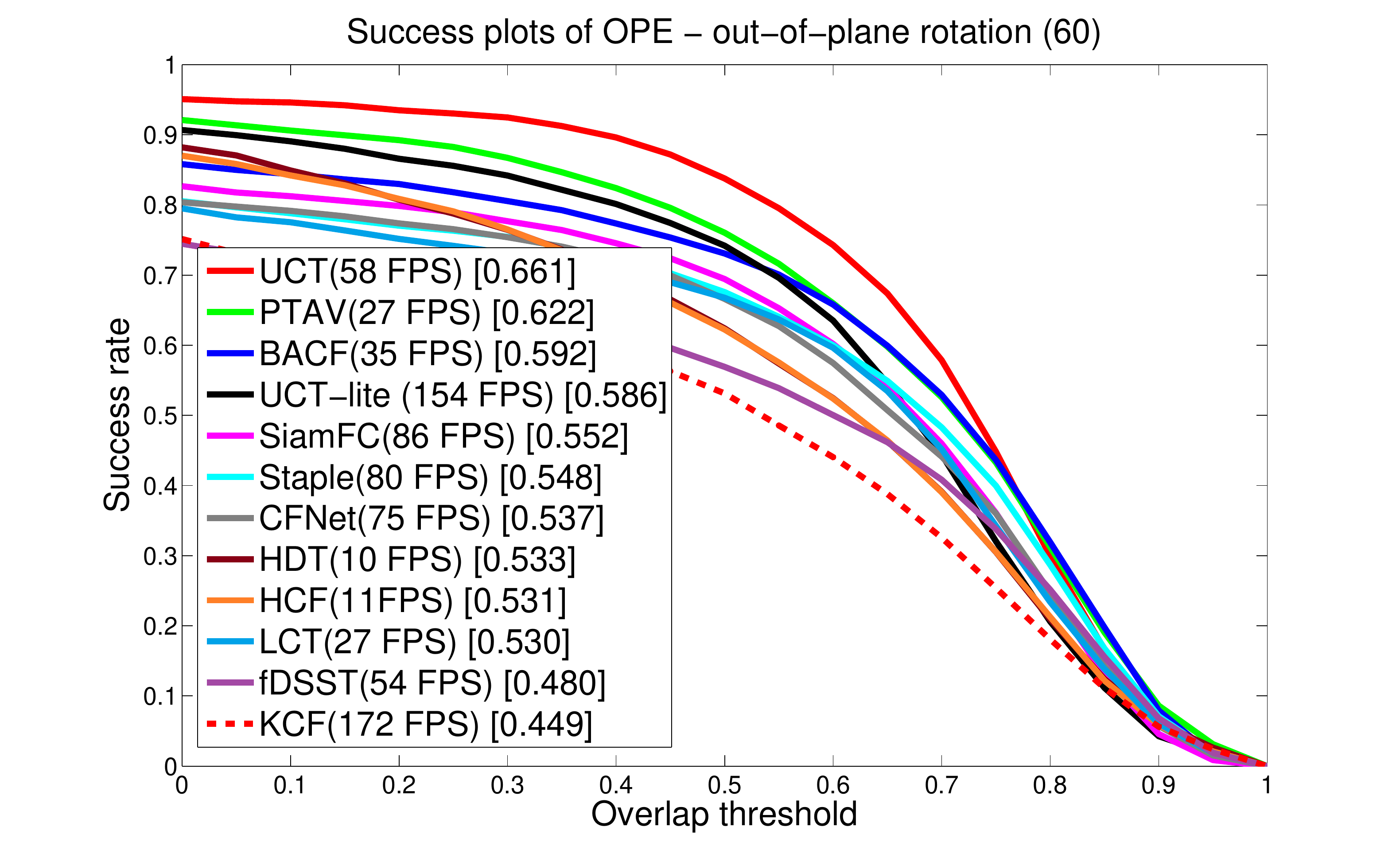}
\end{minipage}%
\begin{minipage}[c]{4.7cm}
\includegraphics[width=5cm]{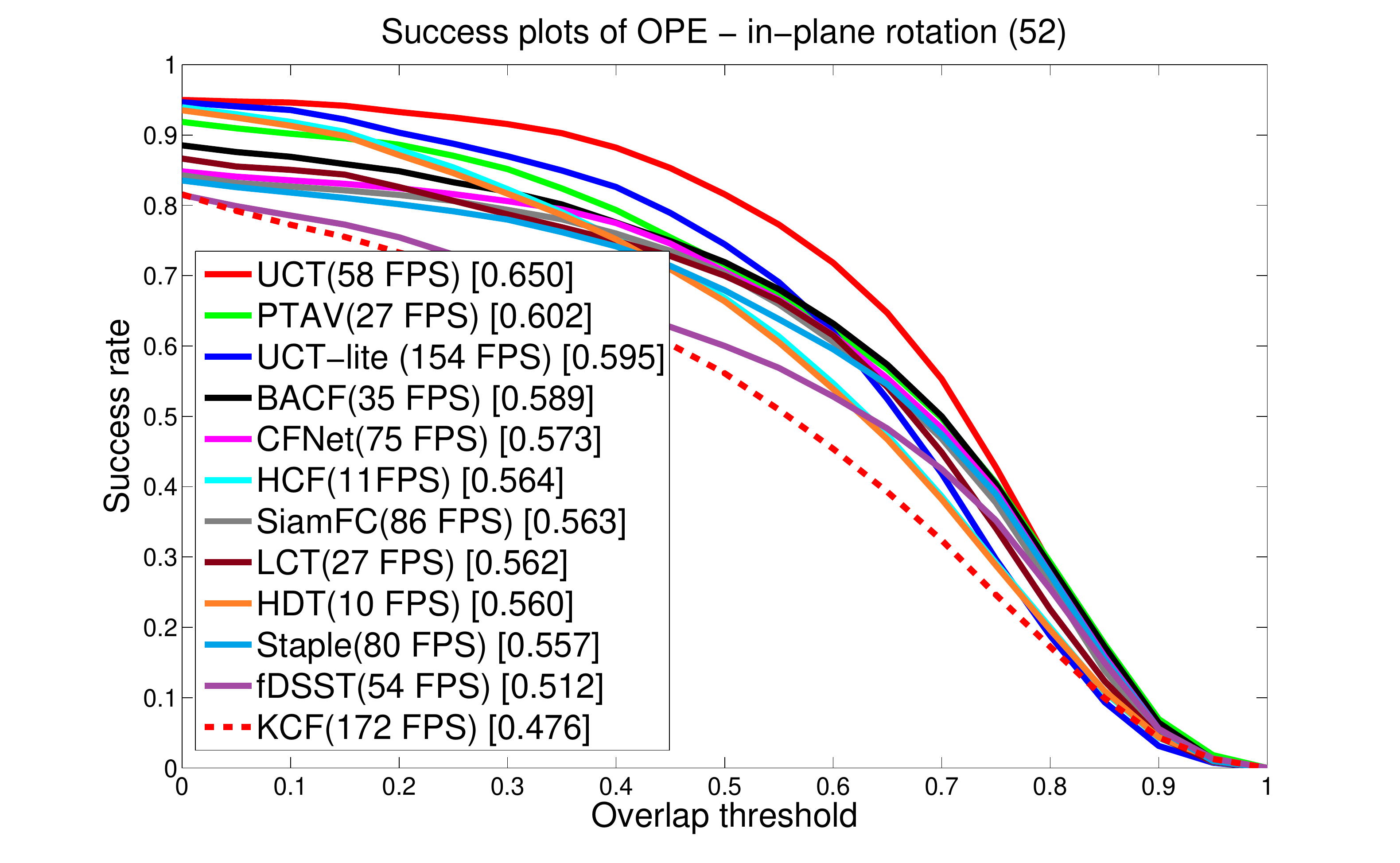}
\end{minipage}%
\begin{minipage}[c]{4.7cm}
\includegraphics[width=5cm]{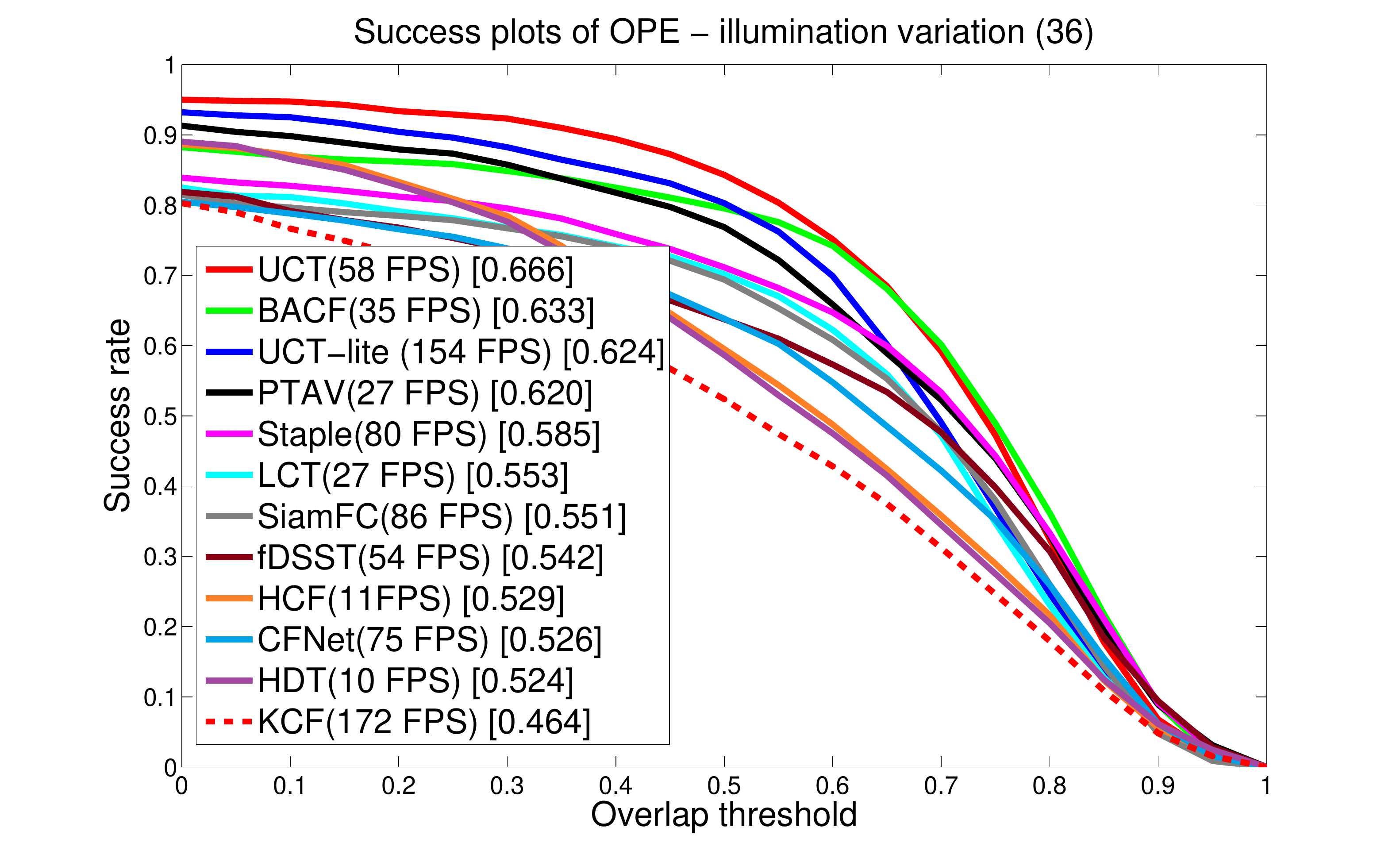}
\end{minipage}%

\begin{minipage}[c]{4.7cm}
\includegraphics[width=5cm]{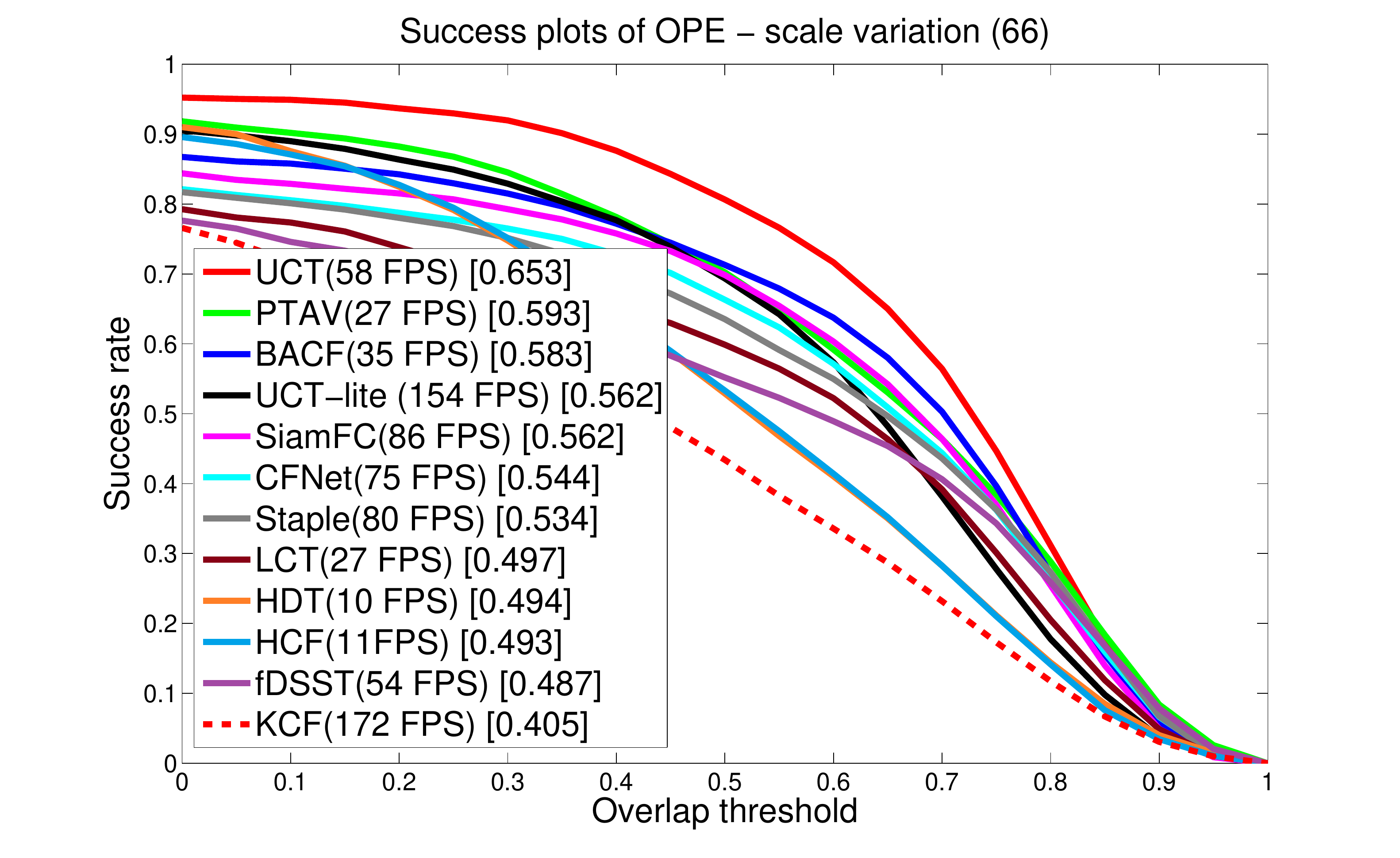}
\end{minipage}%
\begin{minipage}[c]{4.7cm}
\includegraphics[width=5cm]{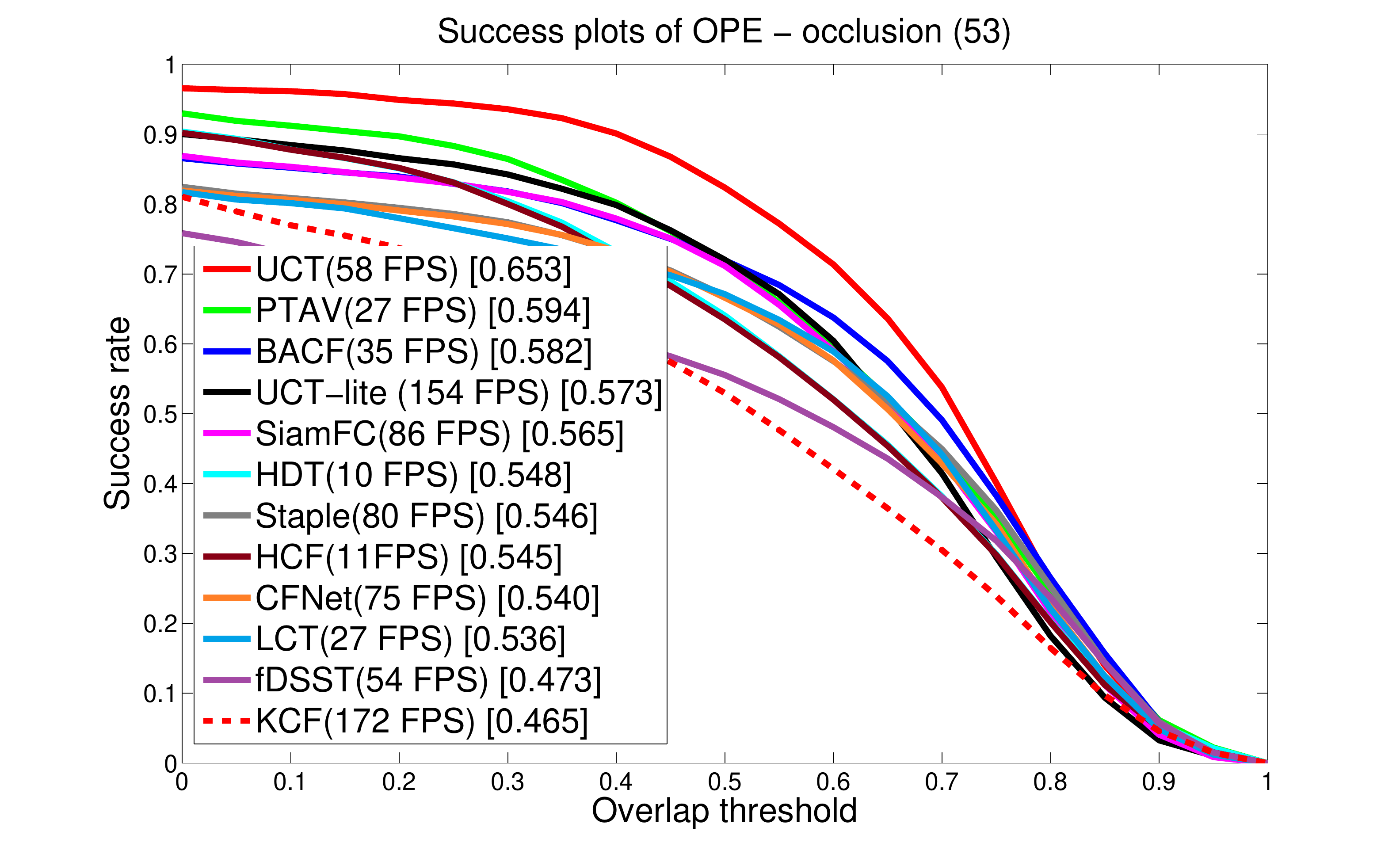}
\end{minipage}%
\begin{minipage}[c]{4.7cm}
\includegraphics[width=5cm]{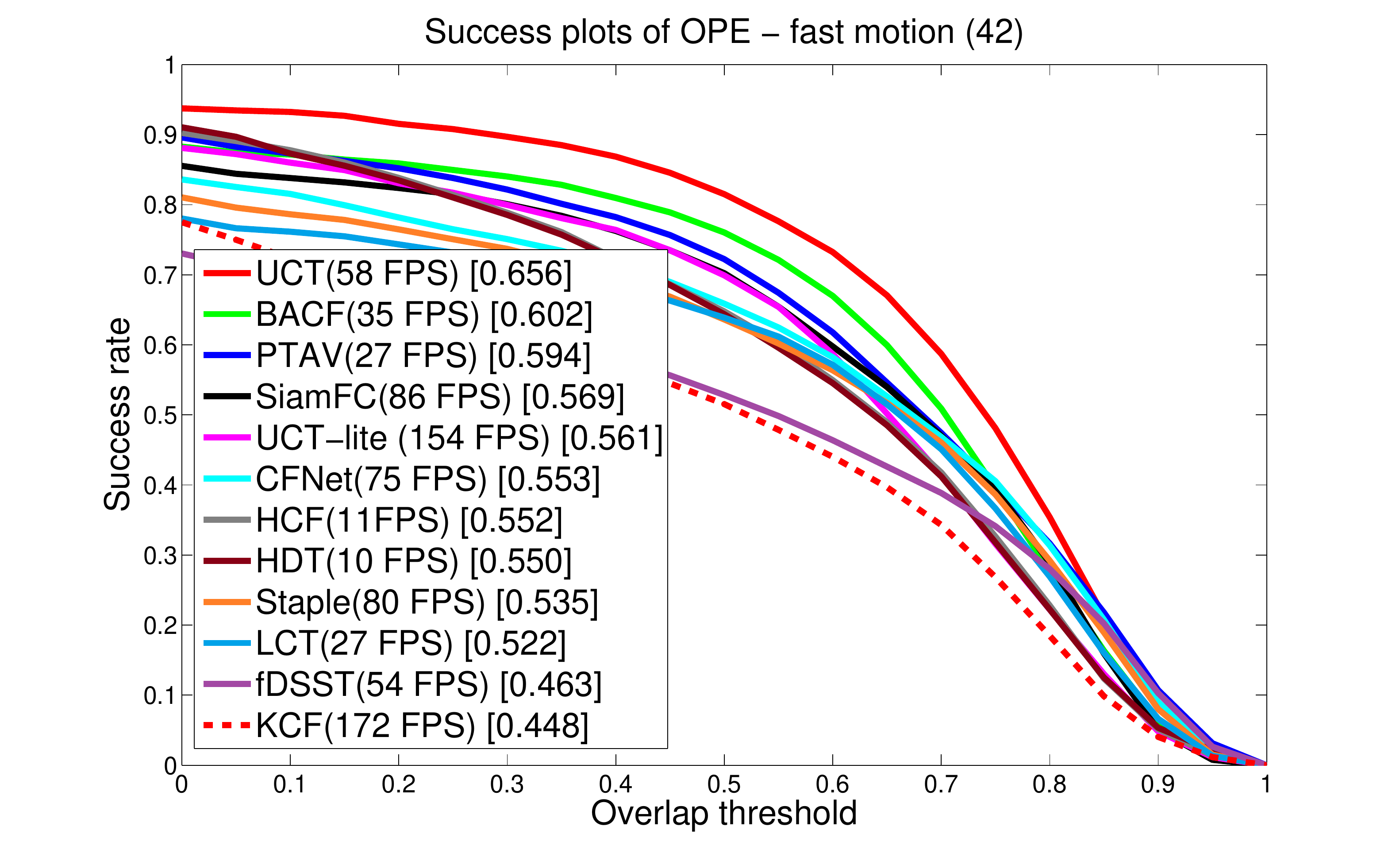}
\end{minipage}%
\begin{minipage}[c]{4.7cm}
\includegraphics[width=5cm]{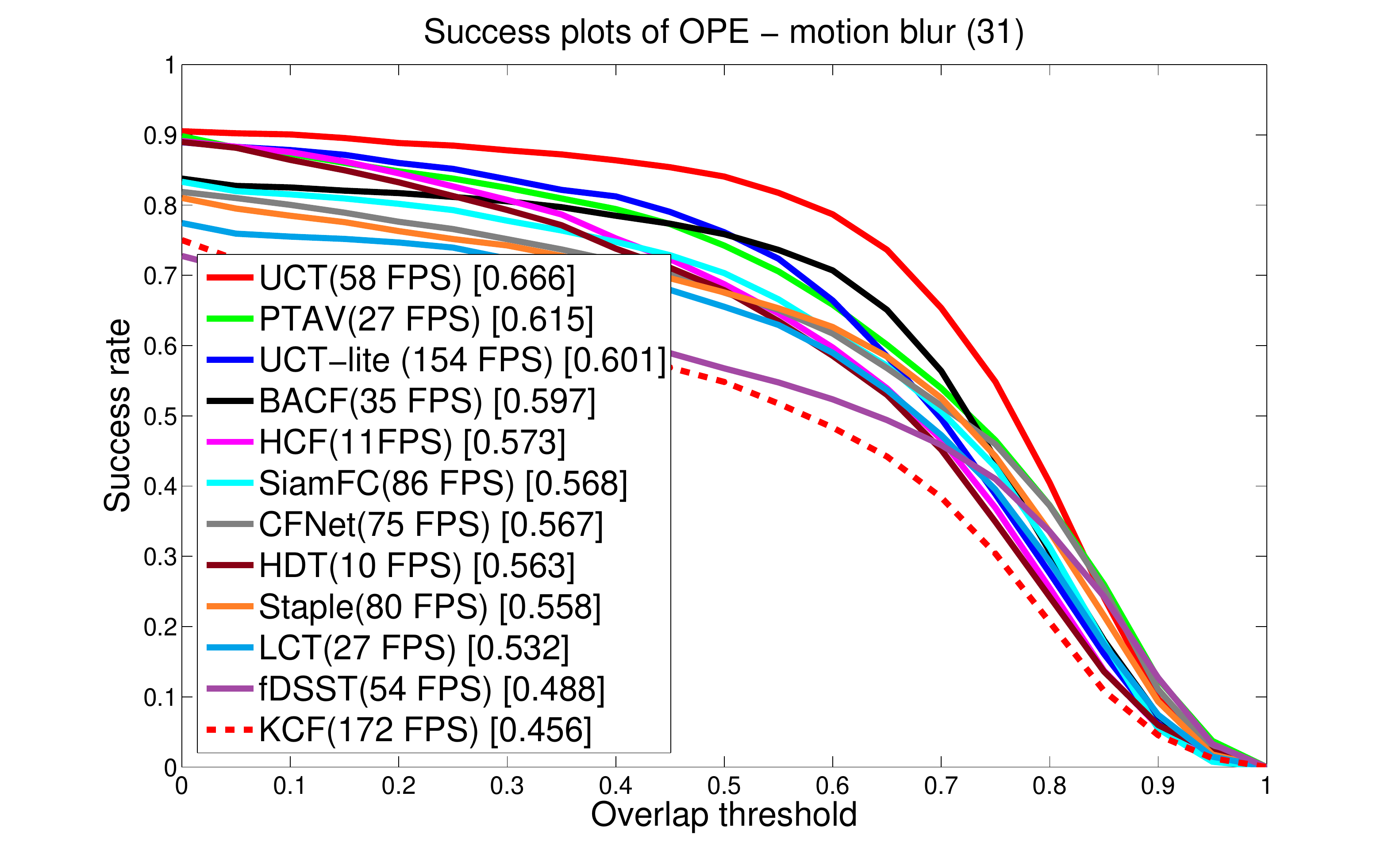}
\end{minipage}%
 \caption{Attributes performance on OTB2015. Best viewed on color display.}
  \label{OTB2015_OPE_attributes}
\end{figure*}

For detailed performance analysis, we report the results on various challenge attributes in OTB2015, such as illumination variation, scale changes, occlusion, etc. Figure~\ref{OTB2015_OPE_attributes} demonstrates that our tracker effectively handles these challenging situations while other trackers obtain lower scores. Comparisons of our approach with four state-of-the-art trackers in the challenging scenarios are shown in Figure~\ref{figure13}.

\begin{table*}[t]
  \centering
  \caption{ Performance on OTB of UCT and compared trackers}
 \begin{tabular}{ccccccc}
    \hline
    \bf Approaches  & \bf Publication & \bf Model update frequency & \bf Scale estimation   & \bf AUC in OTB2013   & \bf AUC in OTB2015 & \bf Speed (FPS) \\
    %\midrule
    \hline
    PTAV & ICCV 2017  & in short-term phase &  yes &0.654 &0.631 &27  \\
    BACF & ICCV 2017  & always &  yes & 0.657 &0.621 &35   \\
    SiamFC &ECCVw 2016  & no &  yes &  0.608&0.582&86  \\
    Staple  &CVPR 2016   & always &  yes&  0.600  &0.581 &80 \\
    CFNet  & CVPR 2017  & always &  yes &  0.611 &0.568 &75 \\
    HDT  & CVPR 2016  & always &  yes & 0.603 &0.564 &10  \\
    HCF  & ICCV 2015  & always &  yes & 0.605  &0.562 &11 \\
    LCT  & CVPR 2015  & always &  yes &  0.628 &0.562 &27 \\
    fDSST  & TPAMI 2017  & always & yes &  0.595 &0.517 &54 \\
    KCF  & TPAMI 2015  & always &  no&  0.514 &0.477 &\bf 172  \\
     \hline
    UCT     &ours& score is satisfied  & yes &  \bf 0.693  &\bf 0.670 & 58   \\
    UCT-lite&ours& score is satisfied    & no  & 0.634 & 0.608 &154     \\
    %\bottomrule
    \hline

  \end{tabular}
   \label{table_all_otb}
\end{table*}
%
%\subsection{Results on VOT}
%
%The Visual Object Tracking (VOT) challenges are well-known competitions in tracking community. The VOT have held several times from 2013 and their results will be reported at ICCV or ECCV. In this subsection, we compare our method, UCT with entries in VOT2015 \cite{c44} and VOT2016 \cite{c10}.

\subsection{Results on VOT2015}

\begin{table}[tp]
\scriptsize
  \centering
  \caption{ Comparisons with top trackers in VOT2015. {\color{red}Red}, {\color{green}green} and {\color{blue}blue} fonts indicate \emph{1st, 2nd, 3rd} performance, respectively. Best viewed on color display.}
\begin{tabular}{cccc}
\hline
\bf Trackers & \bf EAO & \bf Accuracy & \bf Failures  \\
\hline
\textbf{UCT} & \color{red} 0.3576 & \color{red} 0.58 & \color{red} 0.97  \\\hline
\textbf{UCT-lite} & 0.2957 & \color{green} 0.56 & 1.26  \\\hline
\textbf{DeepSRDCF} & \color{green}0.3181 & \color{green}0.56 & \color{blue}1.05  \\\hline
\textbf{EBT}       & \color{blue}0.3130 & 0.47 & \color{green}1.02  \\\hline
\textbf{srdcf}     & 0.2877 & \color{green}0.56 & 1.24  \\\hline
\textbf{LDP}       & 0.2785 & 0.51 & 1.84  \\\hline
\textbf{sPST}      & 0.2767 & \color{blue}0.55 & 1.48   \\\hline
\textbf{scebt}     & 0.2548 & \color{blue}0.55 & 1.86  \\\hline
\textbf{nsamf}     & 0.2536 & 0.53 & 1.29  \\\hline
\textbf{struck}    & 0.2458 & 0.47 & 1.61  \\\hline
\textbf{rajssc}    & 0.2458 & \color{red}0.57 & 1.63   \\\hline
\textbf{s3tracker} & 0.2420 & 0.52 & 1.77 \\\hline
\end{tabular}
   \label{vot2015_table}
\end{table}

VOT2015 \cite{VOT2015} consists of 60 challenging videos that are automatically selected from a 356 sequences pool. The trackers in VOT2015 are evaluated by expected average overlap (EAO) measure, which is the inner product of the empirically estimating the average overlap and the typical-sequence-length distribution. The EAO measures the expected no-reset overlap of a tracker run on a short-term sequence. Besides, accuracy (mean overlap) and robustness (average number of failures) are also reported.

\begin{figure}[tp]
 \centering
\begin{minipage}[c]{8.5cm}
\includegraphics[width=8.5cm]{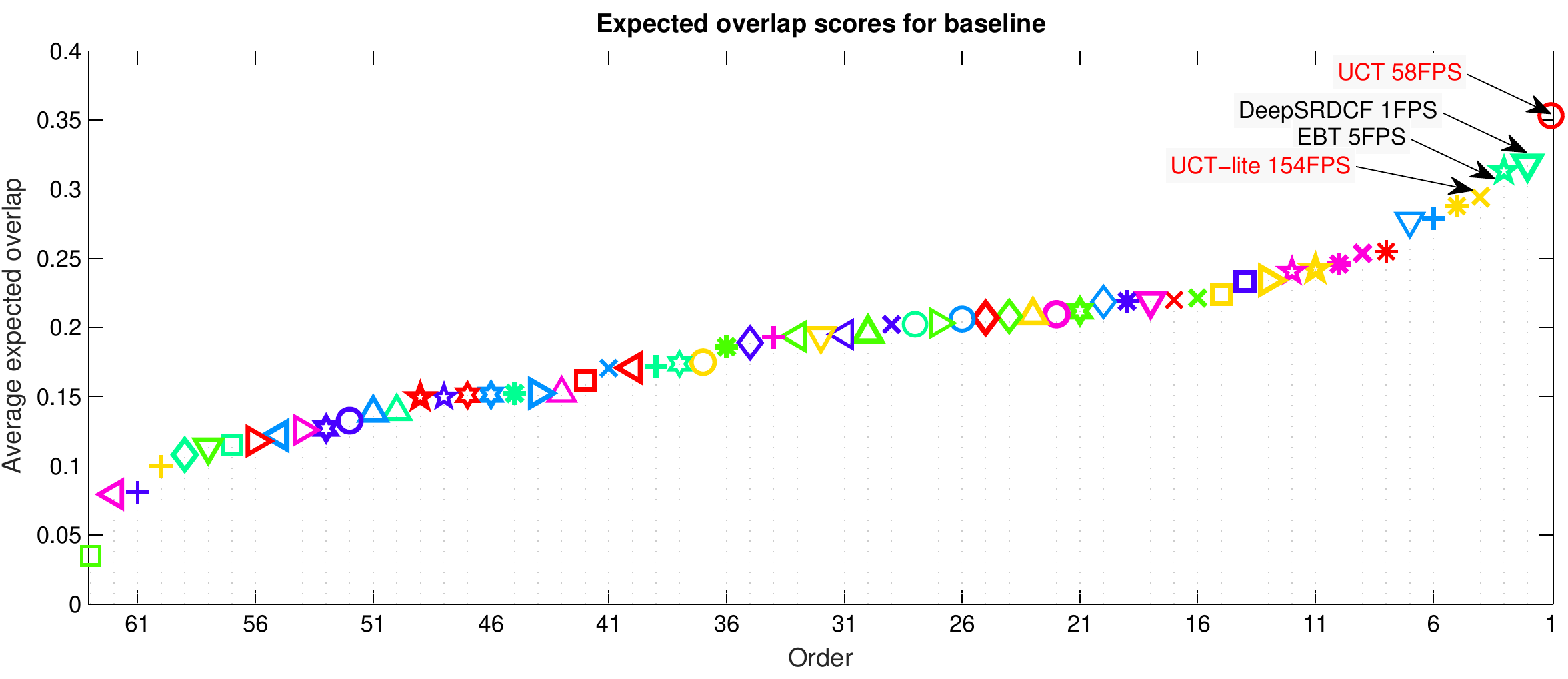}
\end{minipage}%

\begin{minipage}[c]{8.5cm}
\includegraphics[width=8.5cm]{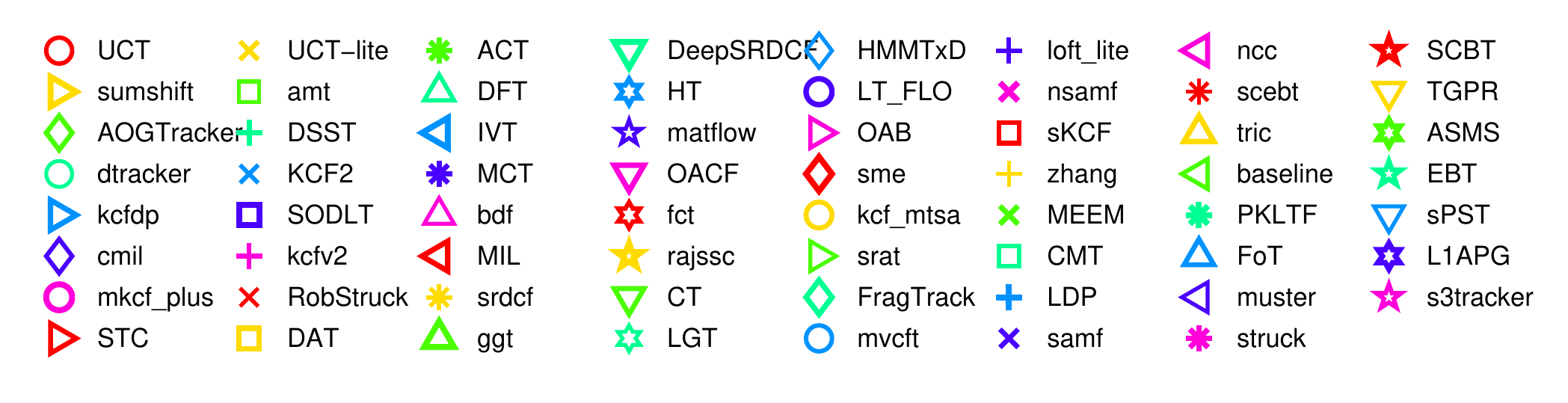}
\end{minipage}%
 \caption{EAO ranking with trackers in VOT2015. The better trackers are located at the right. Best viewed on color display.}
 \label{VOT2015_eao}
\end{figure}
%\begin{figure}[htbp]
%\centering
%\includegraphics[width=0.9\linewidth]{figure12.pdf}
%\caption{Attributes ranking on VOT2015, including camera motion, illumination variation and motion changes. Legends are consistent with Figure ~\ref{figure11}. The better trackers are located at the right.}
%\label{figure12}
%\end{figure}

 %Figure~\ref{VOT2015_eao} illustrates that proposed UCT can rank seventh in EAO measures. None of top six trackers can perform in real-time(their speed is less than 5 EFO). Since UCT employs end-to-end training, efficient updating and scale handling strategies, it can achieve a great balance between performance and speed. For detailed performance analysis, we also report the ranking results on various challenge attributes in VOT2015, including camera motion, illumination variation and motion changes. As shown in Figure~\ref{VOT2015_attributes}, the proposed UCT can rank third, fourth and fifth in three attributes, respectively. It is worth noting that our approach achieves best results compared with other real-time trackers. In Table~\ref{vot2015_table}, we list the EAO, accuracy and failures of FlowTrack and top 10 entries in VOT2015. FlowTrack rank \emph{1st} according to all 3 criterions. The top performance can be attributed to the associating of flow information and end-to-end training framework.

 In VOT2015 experiment, we present a state-of-the-art comparison with the participants in the challenge according to the latest VOT rules (see http://votchallenge.net). It is worth noting that MDNet \cite{MDNet} is not compatible with the latest VOT rules because of OTB training data.

 Figure~\ref{VOT2015_eao} illustrates that our UCT ranks \emph{1st} in 61 trackers according to EAO criterion, while performing at 58 FPS. The faster UCT-lite (154 FPS) can rank \emph{4th} in EAO criterion. In Table~\ref{vot2015_table}, we list the EAO, accuracy and failures of UCT and top 10 entries in VOT2015. UCT ranks \emph{1st} according to all 3 criterions. The top performance can be attributed to the unified convolutional architecture and end-to-end training.

\begin{figure}[thpb]
\setlength{\abovecaptionskip}{0.5cm}
\setlength{\belowcaptionskip}{0.4cm}
  \centering
  \includegraphics[width=1\linewidth]{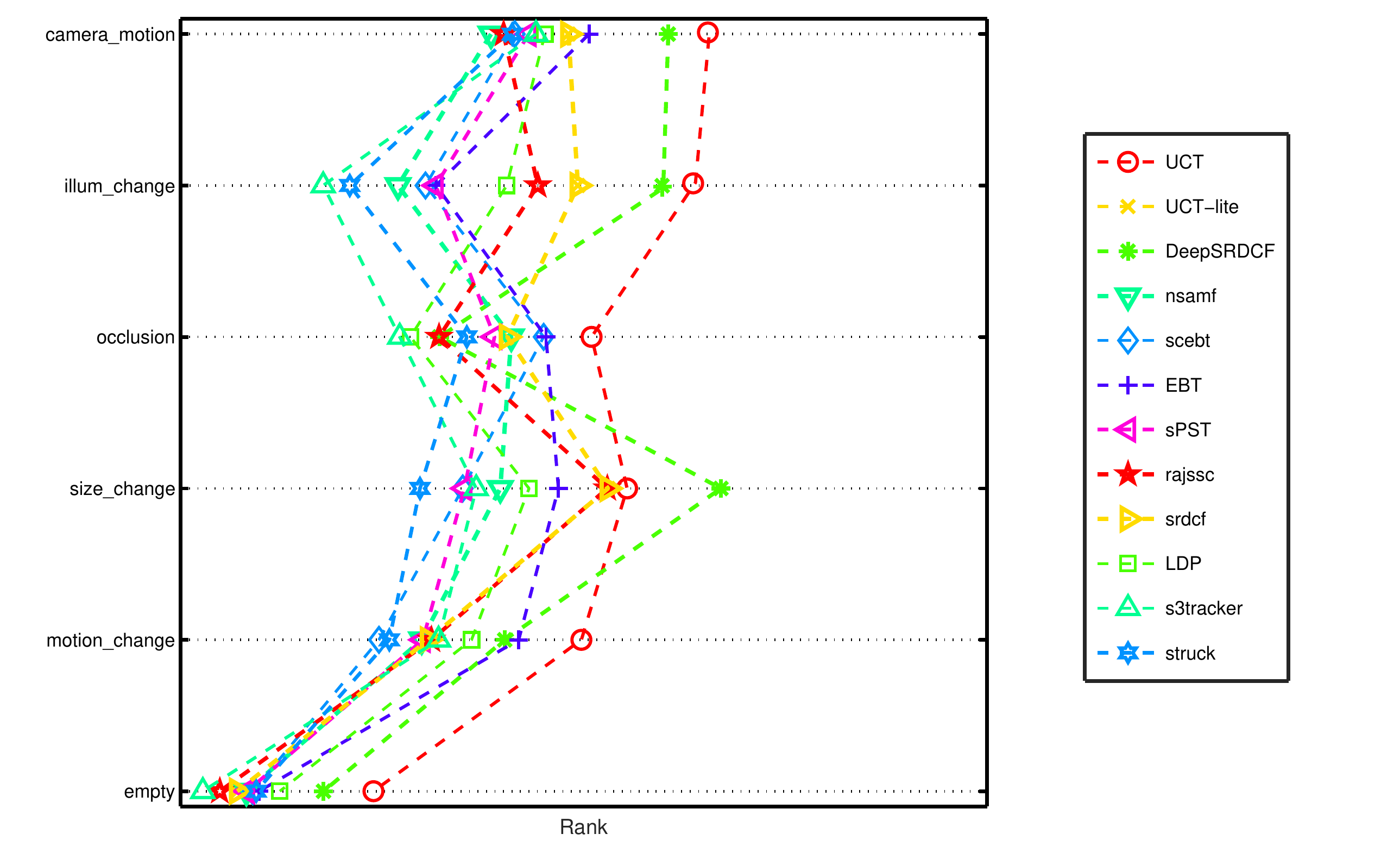}
  \caption{EAO scores for each attribute on the VOT2015 dataset. \emph{empty} denotes frames with no labeled attribute. Best viewed on color display.}
  \label{VOT2015_attributes}
\end{figure}

Additionally, further experimental attributes evaluations on the VOT2015 dataset with 60 videos are presented. In the VOT2015 dataset, each frame is labeled with 5 different attributes: camera motion, illumination change, occlusion, size change and motion change. The performance is evaluated by expected average overlap (EAO) measure. As shown in Figure~\ref{VOT2015_attributes}, our approach (UCT) achieves the best results on 4 in 5 attributes.

\subsection{Results on VOT2016}

The datasets in VOT2016 \cite{VOT2016} are the same as VOT2015, but the ground truth has been re-annotated. VOT2016 also adopts EAO, accuracy and robustness for evaluations.

\begin{figure}[thpb]
 %\centering
\begin{minipage}[c]{8.5cm}
\includegraphics[width=8.5cm]{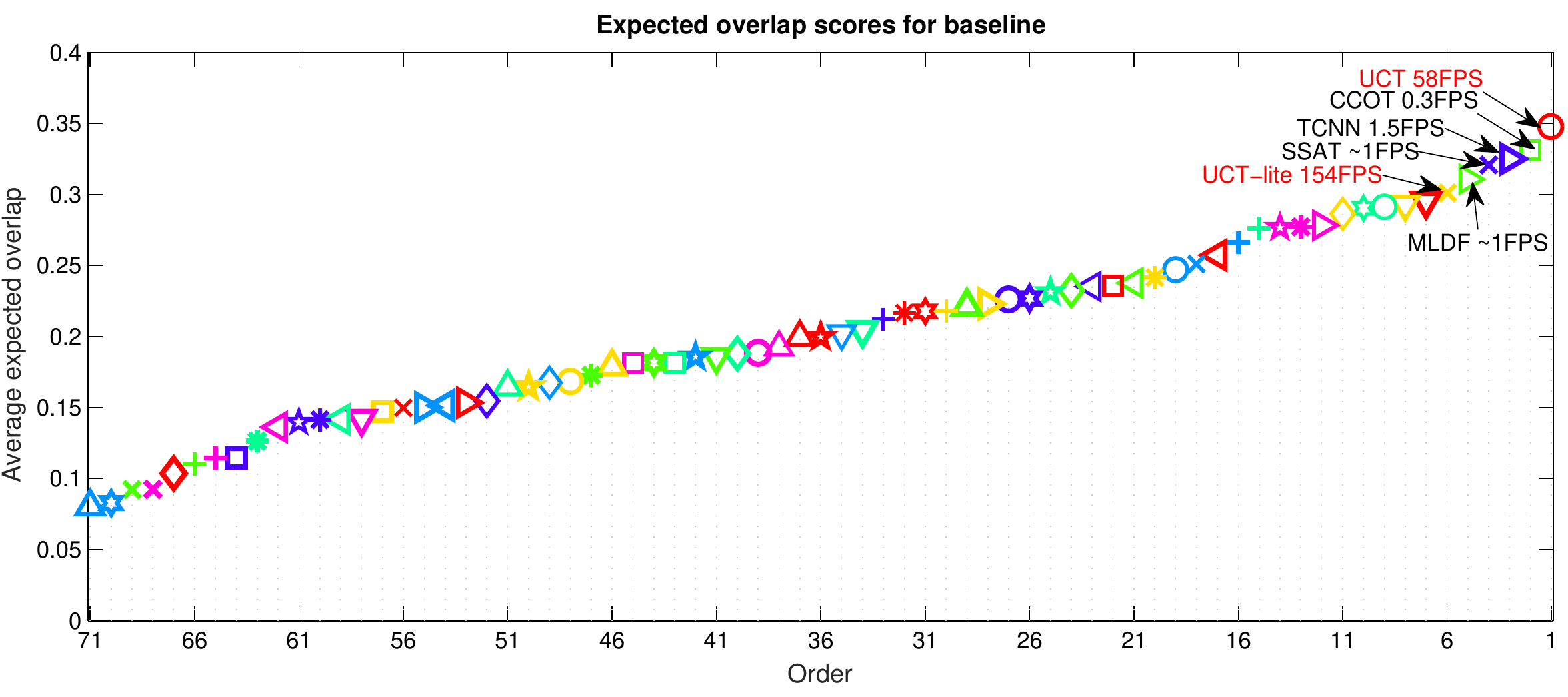}
\end{minipage}%

\begin{minipage}[c]{8.5cm}
\includegraphics[width=8.5cm]{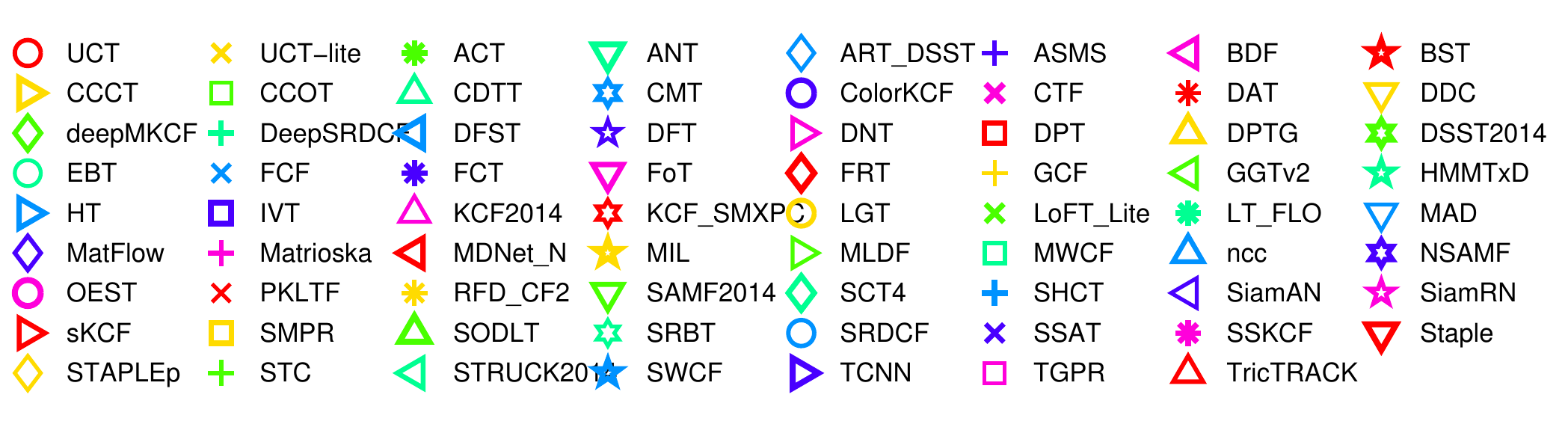}
\end{minipage}%
 \caption{EAO ranking with trackers in VOT2016. The better trackers are located at the right. Best viewed on color display.}
 %\captionsetup[subfloat]{captionskip=0pt,nearskip=0pt,farskip=0pt}
\captionsetup[figure]{position=bottom,belowskip=0pt,aboveskip=0pt}
\captionsetup[table]{belowskip=0pt,aboveskip=0pt}
 \label{VOT2016_eao}
\end{figure}

In experiment, we compare our method with participants in challenges. Figure~\ref{VOT2016_eao} illustrates that our UCT ranks \emph{1st} in 70 trackers according to EAO criterion. It is worth noting that our method can operate at 58 FPS, which is near 200 times faster than CCOT \cite{CCOT} (0.3 FPS). The UCT-lite ranks \emph{6th} with a speed of 154 FPS. Figure~\ref{EAO-EFO} shows the performance and speed of the state-of-the-art trackers. It illustrates that our tracker can achieve a superior performance while operating at high speed.

\begin{figure}[thpb]
\setlength{\abovecaptionskip}{0.5cm}
\setlength{\belowcaptionskip}{0.4cm}
  \centering
  \includegraphics[width=0.9\linewidth]{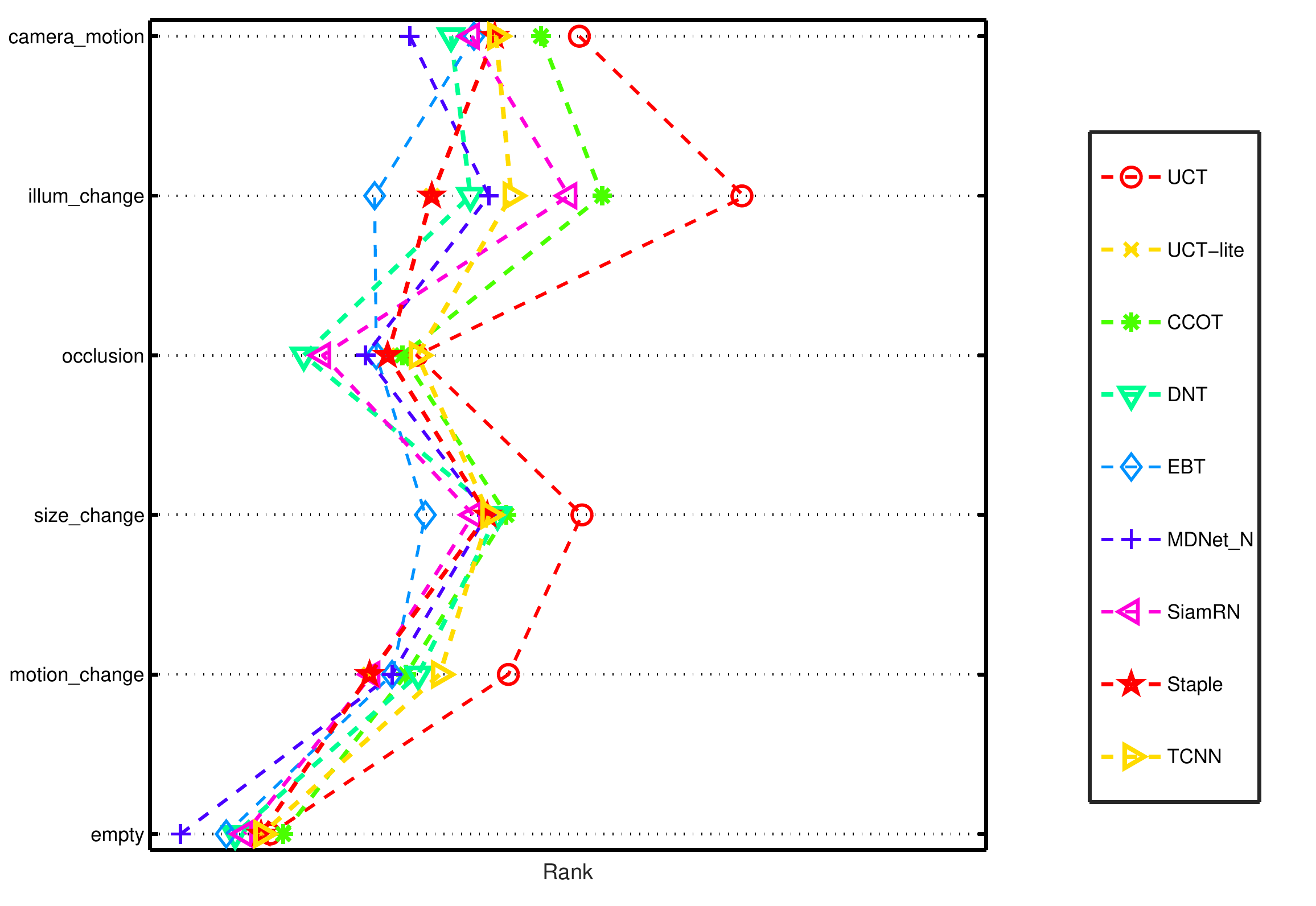}
  \caption{EAO scores for each attribute on the VOT2016 dataset. \emph{empty} denotes frames with no labeled attribute. Best viewed on color display.}
  \label{VOT2016_attributes}
\end{figure}

\begin{figure}[thpb]
 \centering
\begin{minipage}[c]{8cm}
\includegraphics[width=8cm]{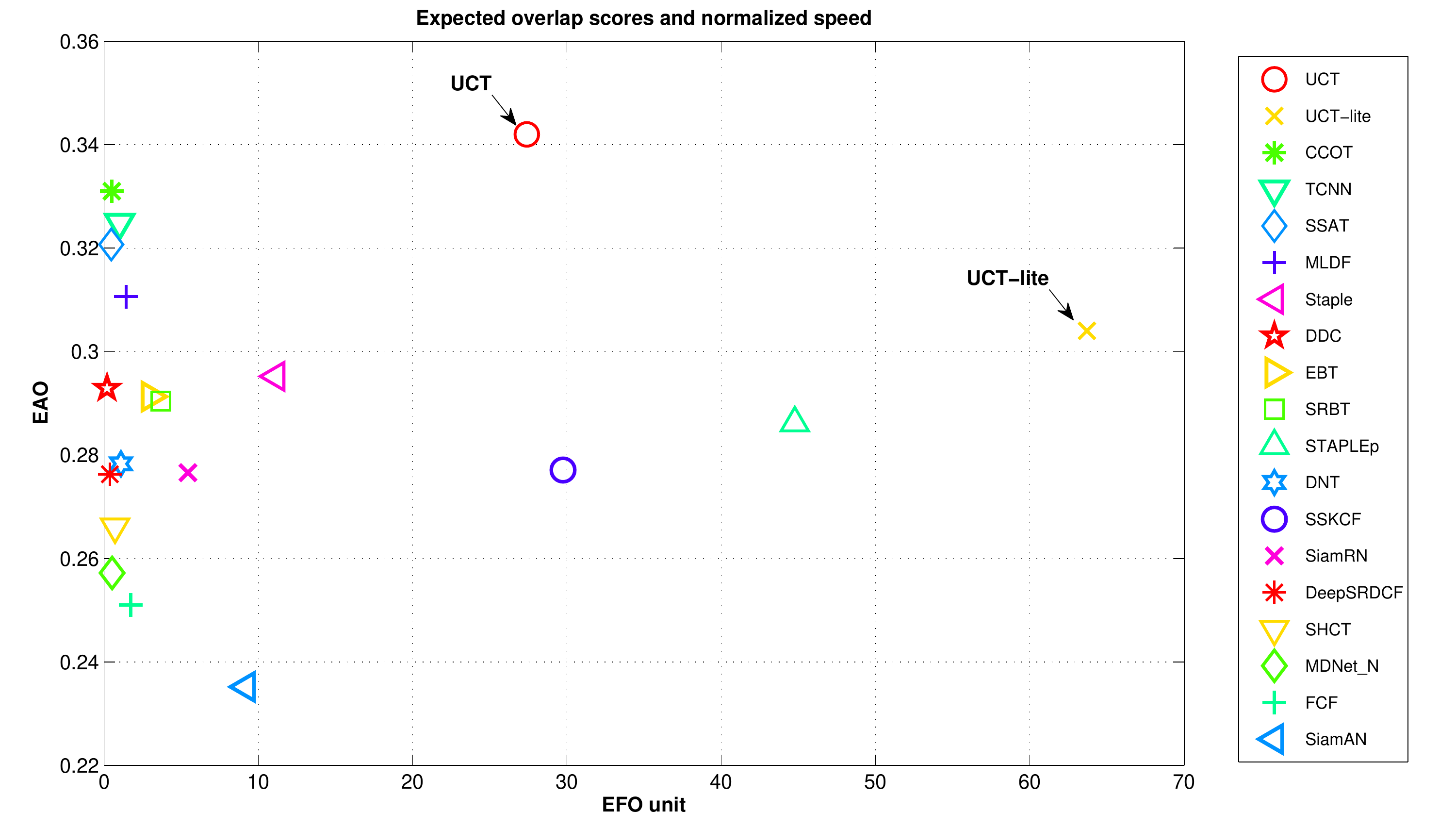}
\end{minipage}%
 \caption{Performance and speed of our tracker and some state-of-the-art trackers in VOT2016. Speed is evaluated by normalized equivalent filter operations  (EFO) \cite{VOT2016}.
 More closed to top means higher precision, and more closed to right means faster. UCT is able to rank \emph{1st} in EAO while operating at 27 EFO (58 FPS). Best viewed on color display.}
 \label{EAO-EFO}
\end{figure}

For detailed performance analysis, we also list accuracy and robustness of representative trackers in VOT2016. As shown in Table~\ref{vot2016_table}, the accuracy and robustness of proposed UCT can rank \emph{1st} and \emph{2nd}, respectively. At last, we provide further experimental attributes evaluation on the VOT2016 dataset with 60 videos. In the VOT2016 dataset, each frame is labeled with five different attributes: camera motion, illumination change, occlusion, size change and motion change. Figure~\ref{VOT2016_attributes} visualizes the EAO of each attribute individually. Our approach (UCT) ranks \emph{1st} in 4 attributes and \emph{2nd} in 1 attributes.
%FlowTrack can rank \emph{2nd} according to EAO criterion, which has a similar performance with CCOT.

\begin{table}[thpb]
\scriptsize
  \centering
  \caption{ Comparisons with top trackers in VOT2016. {\color{red}Red}, {\color{green}green} and {\color{blue}blue} fonts indicate \emph{1st, 2nd, 3rd} performance, respectively. Best viewed on color display.}
\begin{tabular}{cccc}
\hline
\bf Trackers & \bf EAO & \bf Accuracy & \bf Robustness  \\
\hline
\textbf{UCT}& \color{red}0.342   &\color{red}0.569     &\color{green}0.239\\\hline
\textbf{UCT-lite}& 0.304   &\color{blue}0.551     &0.362\\\hline
\textbf{CCOT}     & \color{green}0.331     &0.539                &\color{red}0.238  \\\hline
\textbf{TCNN}     & \color{blue}0.325                &\color{green}0.554   &0.268 \\\hline
\textbf{Staple}   & 0.295    &0.544                 &0.378 \\\hline
\textbf{EBT}      & 0.291                &0.465                &\color{blue}0.252  \\\hline
\textbf{DNT}      & 0.278                &0.515                &0.329  \\\hline
\textbf{SiamFC}   & 0.277                &0.549   &0.382  \\\hline
\textbf{MDNet}    & 0.257                &0.541                &0.337  \\\hline
\end{tabular}
   \label{vot2016_table}
\end{table}

\subsection{Qualitative results}
\begin{figure*}[thpb]
\centering
\includegraphics[width=0.9\linewidth]{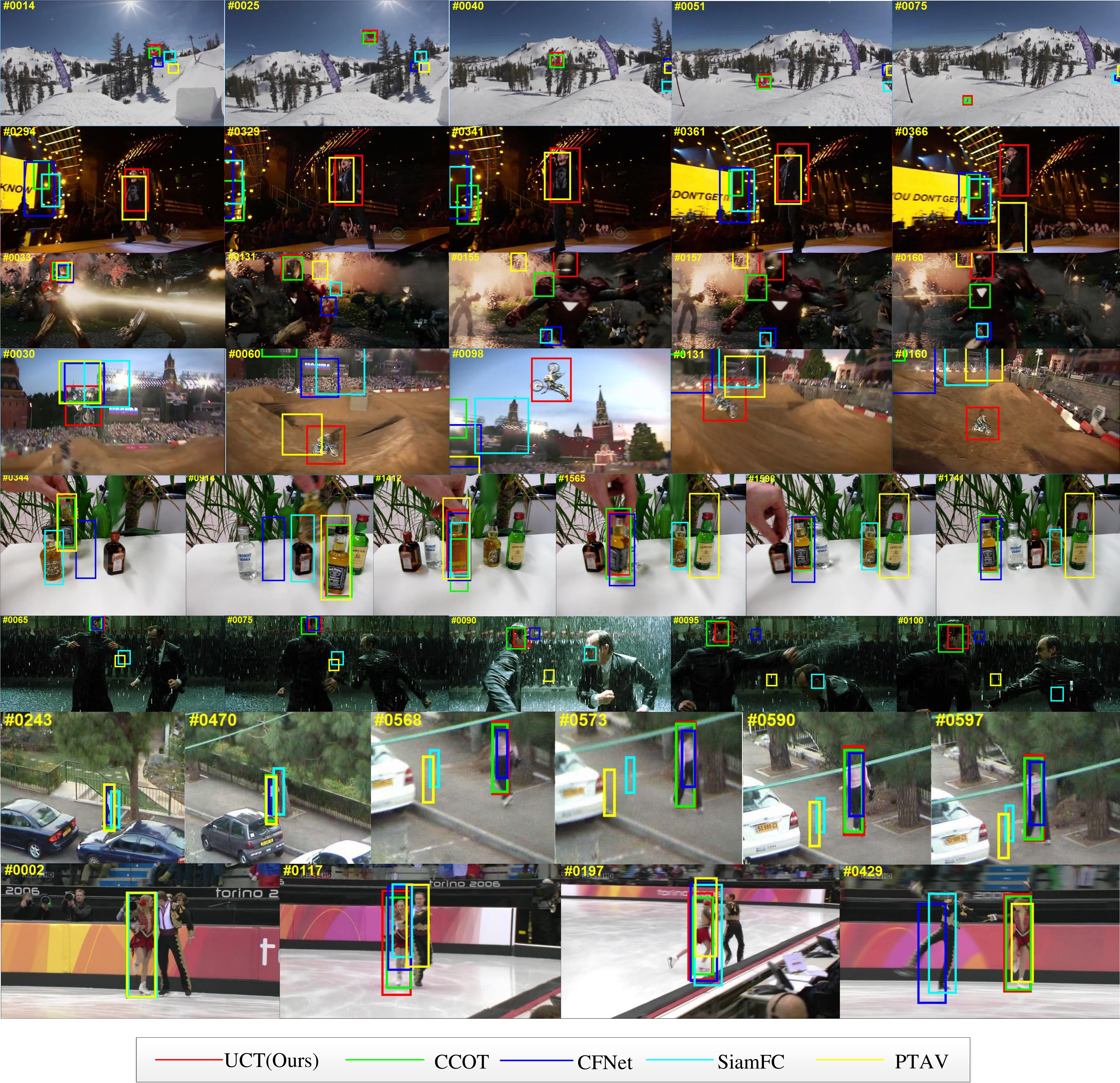}
\caption{Comparisons of our approach with four state-of-the-art trackers in the changing scenario \emph{skiing, singer2, ironman, motorRolling, liquor, matrix, woman, skating2}. The compared trackers are three recent real-time trackers: SiamFC \cite{SiamFC}, CFNet \cite{CFNet}, PTAV \cite{PTAV}, and the winner of VOT2016: CCOT \cite{CCOT}.}
\label{figure13}
\end{figure*}

To visualize the superiority of our framework on tracking performance, we show examples of UCT results compared with recent trackers on challenging sample videos.
As shown in Figure ~\ref{figure13}, the target in sequence $skiing$ undergoes severe fast motion. Only UCT and CCOT successfully tracks until the end, while proposed UCT is more precise than CCOT as shown in \#14, \#25 and \#40. The target in sequence \emph{singer2, ironman} undergoes different severe deformation. Since end-to-end training enables learned CNN features are tightly coupled with tracking process, the proposed UCT can handle these challenges. While CCOT, CFNet, SiamFC and PTAV lead to failure in \#366 of  sequence \emph{singer2} and \#155, \#157, \#160 of sequence \emph{ironman}.
Sequences \emph{liquor} and \emph{matrix} illustrate background clutter challenges, where the UCT results in successful tracking.
Sequences \emph{motorRolling} and \emph{skating2} contain in-plane rotation and out-of-plane rotation. The trained UCT tracks the target while other trackers tend to drift to backgrounds.
Lastly, the proposed UCT handles partial occlusions and scale changes in sequence \emph{woman} such as \#568 and \#597.
%As shown in Figure ~\ref{figure13}, the target in sequence $skiing$ undergoes severe deformation and fast motion. Learned similarity in SiamFC is not discriminative enough in this situation, so the SiamFC tracker loses target and switches to distractor in frame \#36. FCNT employs CNN features that pre-trains in different task,leading to substantial drifting in frame \#309. In contrast, the proposed UCT adopts end-to-end training and efficient tracking strategy, resulting in successful tracking in this sequence. $Soccer$ is a sequences that with attributes of occlusion and pose variations, only proposed method can handle these challenges while other trackers drift to background. In sequence $jogging2$, the target undergoes occlusion and distractor. The compared FCNT and Staple fail to recover from occlusion as shown in frame \#61 and \#74, while SiamFC drifts to distractor in frame \#301. Only proposed UCT success in recover from occlusion and handle distractor successfully.

\section{Conclusions}
In this work, we propose a high performance unified convolutional tracker (UCT) that learn the convolutional features and perform the tracking process simultaneously. In online tracking, efficient updating and scale handling strategies are incorporated into the network. It is worth emphasizing that our proposed algorithm not only performs superiorly, but also runs at a very fast speed which is significant for real-time applications. On VOT2016 dataset, the proposed UCT obtains an EAO of 0.342 and performs at 58 FPS, yielding 3.2\% relative gain in performance and 200 times gain in speed compared with challenge winner-CCOT~\cite{CCOT}.

Future directions of this paper will try to apply the proposed UCT framework to mobile robot with active vision systems, and it would be interesting to explore the more efficient network architecture such as MobileNet and ShuffleNet.
%Experiments are performed OTB2013, OTB2015, VOT2015 and VOT2016, and our method achieves leading performance with high speed.

\section*{Acknowledgment}

This work is supported in part by the National Natural Science Foundation of China under Grant No. 61403378 and 51405484, and in part by the National High Technology Research and Development Program of China under Grant No.2015AA042307.

%------------------------------------------------------------------------

%{\small
%\bibliographystyle{ieee}
%\bibliography{egbib}
%}

\bibliographystyle{IEEEtran}
\bibliography{IEEEabrv,egbib}

\end{document}